\documentclass[letterpaper]{article} % DO NOT CHANGE THIS
\usepackage[arxiv]{aaai23}  % DO NOT CHANGE THIS
\usepackage{times}  % DO NOT CHANGE THIS
\usepackage{helvet}  % DO NOT CHANGE THIS
\usepackage{courier}  % DO NOT CHANGE THIS
\usepackage[hyphens]{url}  % DO NOT CHANGE THIS
\usepackage{graphicx} % DO NOT CHANGE THIS
\usepackage{natbib}  % DO NOT CHANGE THIS AND DO NOT ADD ANY OPTIONS TO IT
\usepackage{caption} % DO NOT CHANGE THIS AND DO NOT ADD ANY OPTIONS TO IT
\usepackage{algorithm}
\usepackage{algorithmic}

\usepackage{amsfonts,bm}
\usepackage{amsmath,amssymb} % define this before the line numbering.
\usepackage{color}
\usepackage{multirow}
\usepackage[T1]{fontenc}
\usepackage{newfloat}
\usepackage{listings}
\DeclareCaptionStyle{ruled}{labelfont=normalfont,labelsep=colon,strut=off} % DO NOT CHANGE THIS
\floatstyle{ruled}
\newfloat{listing}{tb}{lst}{}
\floatname{listing}{Listing}
\usepackage{colortbl}
\definecolor{mygray}{gray}{.9}
\definecolor{mypink}{rgb}{.99,.91,.95}
\usepackage{xspace}
\usepackage{pifont}
\newcommand{\cmark}{\ding{51}\xspace}%
\newcommand{\xmark}{\ding{55}\xspace}%

\title{NLIP: Noise-robust Language-Image Pre-training}
\author {
    Runhui Huang,\textsuperscript{\rm 1 \equalcontrib}
    Yanxin Long,\textsuperscript{\rm 1 \equalcontrib}
    Jianhua Han,\textsuperscript{\rm 2}
    Hang Xu,\textsuperscript{\rm 2} \\
    Xiwen Liang,\textsuperscript{\rm 1}
    Chunjing Xu,\textsuperscript{\rm 2}
    Xiaodan Liang,\textsuperscript{\rm 1 \dag\thanks{Corresponding author}}
}
\affiliations {
    \textsuperscript{\rm 1} Shenzhen campus of Sun Yat-sen University,
    \textsuperscript{\rm 2} Huawei Noah's Ark Lab\\
    \{huangrh9, longyx9\}@mail2.sysu.edu.cn, hanjianhua4@huawei.com, chromexbjxh@gmail.com, \\ liangcici5@gmail.com, xuchunjing@huawei.com, xdliang328@gmail.com
}
\begin{document}

\maketitle

\begin{abstract}
Large-scale cross-modal pre-training paradigms have recently shown ubiquitous success on a wide range of downstream tasks, e.g., zero-shot classification, retrieval and image captioning. However, their successes highly rely on the scale and quality of web-crawled data that naturally contain much incomplete and noisy information (e.g., wrong or irrelevant content). Existing works either design manual rules to clean data or generate pseudo-targets as auxiliary signals for reducing noise impact, which do not explicitly tackle both the \emph{incorrect} and \emph{incomplete} challenges at the same time. In this paper, to automatically mitigate the impact of noise by solely mining over existing data, we propose a principled \textbf{N}oise-robust \textbf{L}anguage-\textbf{I}mage \textbf{P}re-training framework~(\textbf{NLIP}) to stabilize pre-training via two schemes: \emph{noise-harmonization} and \emph{noise-completion}. First, in \emph{noise-harmonization} scheme, NLIP estimates the noise probability of each pair according to the memorization effect of cross-modal transformers, then adopts noise-adaptive regularization to harmonize the cross-modal alignments with varying degrees. Second, in \emph{noise-completion} scheme, to enrich the missing object information of text, NLIP injects a concept-conditioned cross-modal decoder to obtain semantic-consistent synthetic captions to complete noisy ones, which uses the retrieved visual concepts~(\textit{i.e.}, objects' names) for the corresponding image to guide captioning generation. By collaboratively optimizing noise-harmonization and noise-completion schemes, our NLIP can alleviate the common noise effects during image-text pre-training in a more efficient way. Extensive experiments show the significant performance improvements of our NLIP using only 26M data over existing pre-trained models (e.g., CLIP, BLIP) on 12 zero-shot classification datasets ~(e.g., +8.6\% over CLIP on average accuracy), MSCOCO image captioning~(e.g., +1.9 over BLIP trained with 129M data on CIDEr) and zero-shot image-text retrieval tasks. 
\end{abstract}

\section{Introduction}
\begin{figure}[t!]
		\centering
\includegraphics[width=\linewidth]{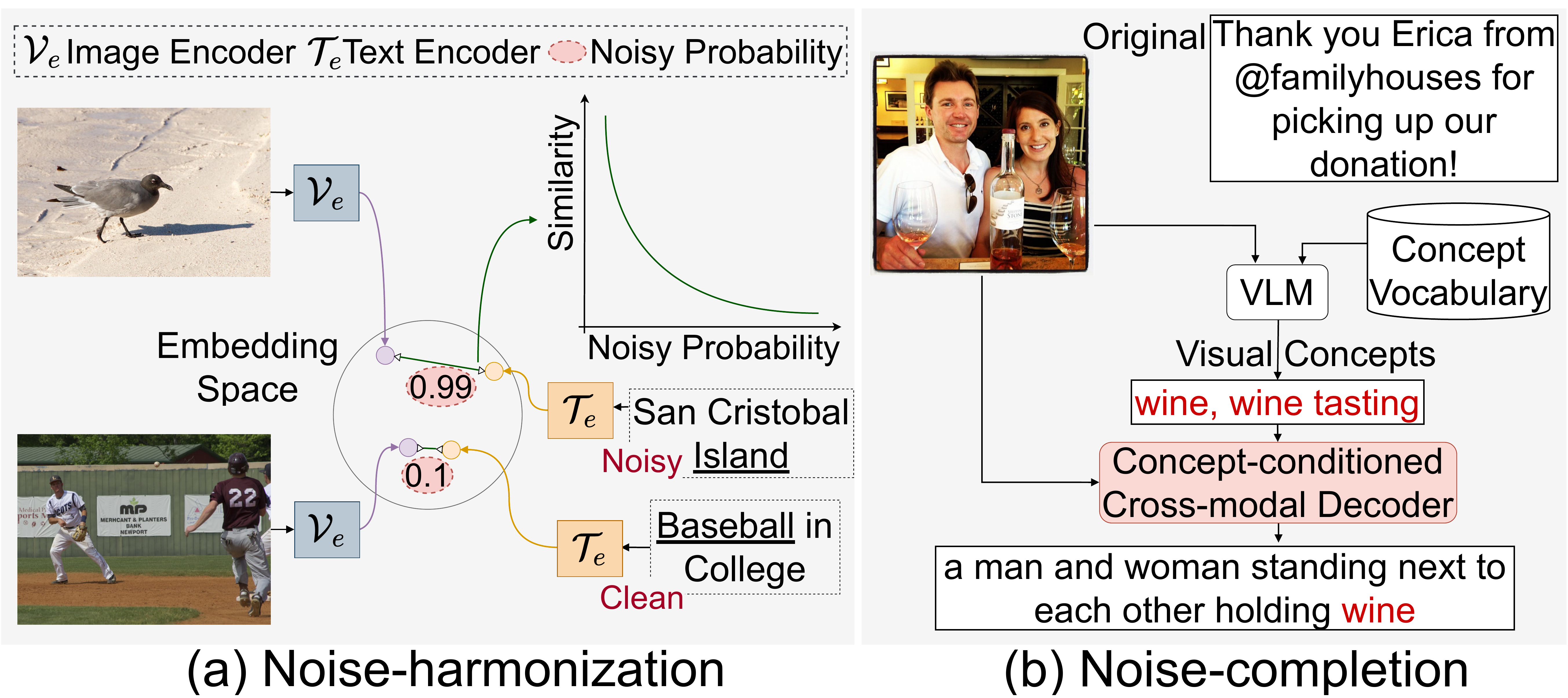}
		\caption{Illustration of 
		two proposed schemes.
		(a) \emph{Noise-harmonization}: NLIP estimates the noise probability of each image-text pair and enforces the pairs with larger noise probability to have fewer similarities in embedding space. (b) \emph{Noise-completion}: NLIP generates enriched descriptions via a concept-conditioned captioner by taking visual concepts retrieved from a vocabulary as auxiliary inputs.}
		\label{fig:introduction}
\end{figure}
Vision-Language Models~(VLMs)~\cite{yao2021filip,radford2021learning,li2021align,jia2021scaling,li2022blip} pre-trained with image-text pairs has shown its extraordinary zero-shot transfer abilities in different downstream tasks, including zero-shot classification~\cite{radford2021learning,yao2021filip}, image-text retrieval \cite{radford2021learning,yao2021filip}, image captioning~\cite{wang2021simvlm} and text-to-image generation~\cite{patashnik2021styleclip}, etc.
Previous works~\cite{radford2021learning,li2022blip} show that the downstream performance of VLMs highly relies on the scale or the quality of pre-training image-caption pairs.
However, considering the prohibitive expense of acquiring high-quality annotated image-caption datasets~\cite{lin2014microsoft}, current paradigms resort to collecting increasingly larger sizes of unlabeled image-text datasets~\cite{thomee2016yfcc100m,sharma2018conceptual}, largely overlooking the prevalent noise in the web. They thus lead to the heavier computation burden and make the pre-training process severely unstable due to the negative impact of noise.

To leverage the advantages of both quality and scale, several attempts have been made to mitigate the negative impact of noisy pairs. On the one hand, some filtering and post-processing procedures~\cite{sharma2018conceptual,changpinyo2021cc12m,jia2021scaling} have been designed to clean up the large-scale unlabeled data for pre-training. On the other hand, few works explore automatic ways during training. For example, ALBEF~\cite{li2021align} resorts to a momentum model to generate pseudo-targets as additional supervision. BLIP~\cite{li2022blip} uses a filter to remove the noisy data rectified by the similarity of image-text pairs and a captioner to regenerate texts. NCR~\cite{huang2021learning} utilizes the loss distribution to divide clean samples and noisy samples and then rectify the labels by model predictions. However, unlabeled ``noise" data often naturally appear with either \textbf{incorrect} text descriptions or \textbf{incomplete} ones (\emph{e.g.}, missing descriptions of some object concepts), where none of the existing works consider automatically alleviating both of them within one framework. Here, we aim to achieve noise-robust learning from two aspects: self-diagnosing incorrect vs. correct pairs and harmonizing the loss; self-generating and selecting confident captions with enriched concepts.

To fully utilize the entire image-caption pairs including the noisy ones, we introduce a principled \textbf{N}oise-robust \textbf{L}anguage-\textbf{I}mage \textbf{P}re-training framework (\textbf{NLIP}) to stabilize pre-training by \emph{noise-harmonization} and \emph{noise-completion} schemes: (a) \textbf{Noise-harmonization}, where NLIP learns to harmonize the cross-modal alignment and adopts noise-adaptive regularization for each pair based on the estimated noisy probability. Specifically, \citet{arpit2017closer} suggests that deep network tends to fit the easy~(\textit{i.e.}, clean) samples first and then the noisy ones. Based on the memorization effect of cross-modal transformers, NLIP first estimates the noise probability for each pair, then applies a noise-adaptive regularization on the image-text contrastive loss to avoid over-fitting to the noisy data~(shown in Fig.\ref{fig:introduction}(a)). This scheme pulls the embeddings of the image and caption in the clean pair more tightly than the one with a higher noisy probability. (b) \textbf{Noise-completion}, where NLIP employs a concept-conditioned cross-modal decoder to synthesize semantic-consistent captions to replace the detrimental noisy texts. Specifically, to guide the caption generation procedure via providing prior information about the existing objects, we first retrieve the visual concepts (\emph{i.e.}, names of existing objects) for each image via a pre-trained VLM. Then these visual concepts and the image are fed into an additional caption head to generate the enriched descriptions for each noisy pair to substitute the noisy caption~(shown in Fig.\ref{fig:introduction}(b)). Furthermore, inspired by~\citet{he2021masked}, we further explore enhancing the visual encoder via randomly masking the input image tokens and then reconstructing them, which can help reduce the computation cost during training and boost visual embedding by maintaining low-level visual information.

Experimental results show that NLIP achieves significant performance on several downstream tasks, including zero-shot classification, zero-shot image-to-text/text-to-image retrieval and image-captioning tasks. Our NLIP outperforms CLIP~\cite{radford2021learning} by 8.6\% in terms of average accuracy on 12 zero-shot classification datasets. With respect to image captioning, NLIP is superior to existing image captioning methods that are trained with substantially more data, \emph{e.g.}, 1.9 over BLIP~\cite{li2022blip} trained with 129M image-text pairs in terms of CIDEr on MSCOCO. For zero-shot image-text retrieval tasks, NLIP surpasses CLIP by 28.7\% in terms of R@1 on Flickr30k. 

\section{Related Work}
\textbf{Vision Language Pre-training (VLP)} models recently garner increasing attention as the surprisingly superior performances on diverse zero-shot downstream tasks. They propose to learn semantic alignments across image and language modalities by pre-training on large-scale data which brings strong performance benefits in downstream tasks (\emph{e.g.}, zero-shot classification, zero-shot retrieval, image caption). Existing VLP models often appear with either encoder-only or encoder-decoder architectures. The encoder-only architectures~\cite{radford2021learning,jia2021scaling,yao2021filip,yuan2021florence,mu2021slip,li2022supervision,you2022learning} aim to align the visual features with textual features in a common cross-modal  semantic space. The encoder-decoder architectures~\cite{wang2021simvlm,li2022blip} employ autoregressive Language Modeling (LM) (\emph{e.g.}, image captioning, text-grounded image generation) to supervise the decoder and excel in generation-related downstream tasks. 
Despite the nature merits in data diversity, the large-scale web-crawled image-text pairs contain much noise (\textit{i.e.}, incomplete or even error information)~\cite{thomee2016yfcc100m,changpinyo2021cc12m}. Some works attempt to mitigate the impact in two aspects. From the data perspective, some strict rules are used to clean up the data~\cite{sharma2018conceptual,changpinyo2021cc12m,jia2021scaling}. From the modeling perspective, ALBEF~\cite{li2021align} adopts momentum models to generate pseudo-targets as additional supervision; BLIP~\cite{li2022blip} presents a filter to remove the noisy data rectified by the similarity of image-text pairs and a captioner to regenerate the corresponding web texts. However, they have not explicitly stabilized and harmonized the pre-training objectives by reevaluating noisy data in a soft way. In this work, we alleviate the noisy impact by simultaneously addressing incorrect and incomplete image-text pairs. Two novel noise-harmonization and noise-completion schemes are collaborative to achieve noise-robust pre-training. 
\begin{figure*}[t!]
		\begin{center}
        \includegraphics[width=0.95\linewidth]{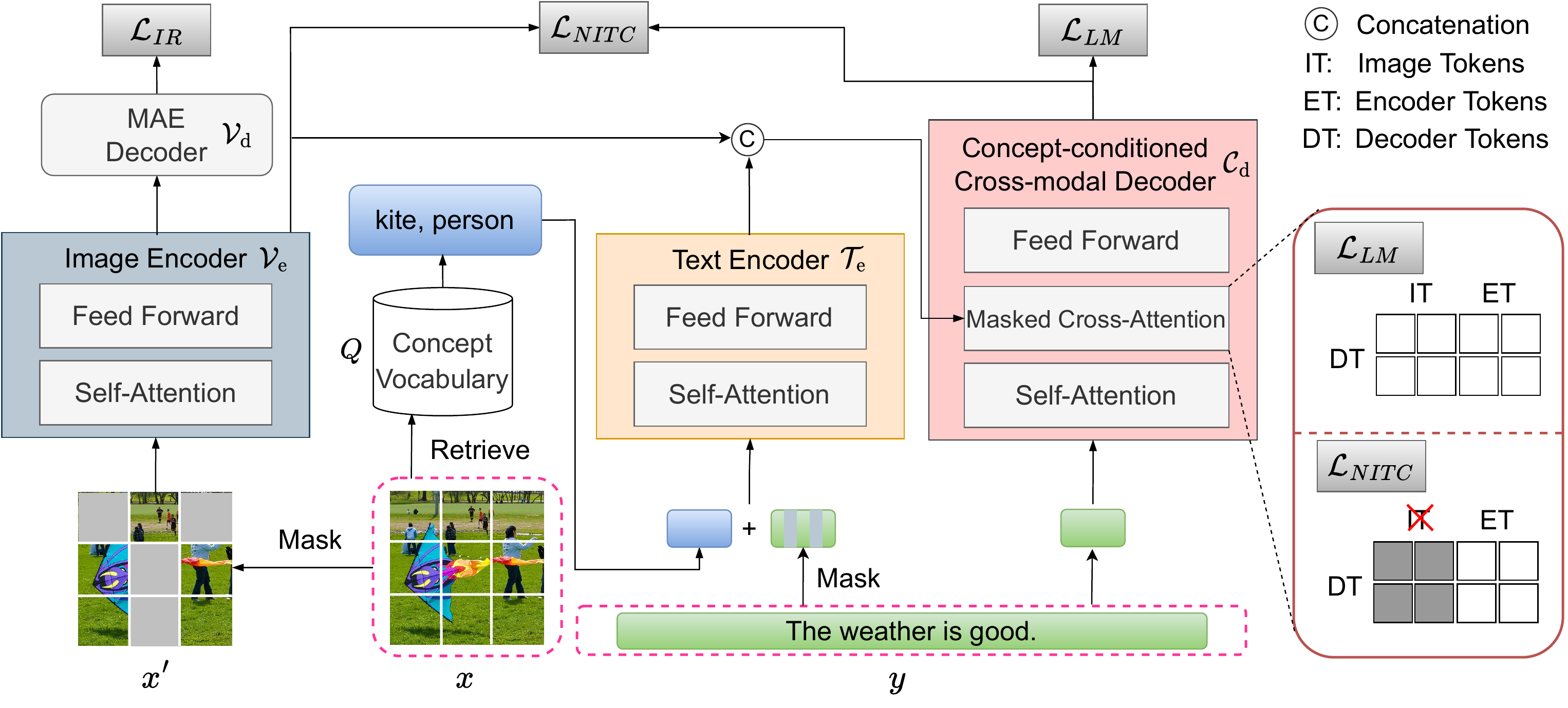}
		\end{center}
		\caption{Overview of the proposed NLIP architecture. NLIP consists of an image encoder $\mathcal{V}_{\rm e}$, text encoder $\mathcal{T}_{\rm e}$, cross-modal decoder $\mathcal{C}_{\rm d}$ and MAE decoder $\mathcal{V}_{\rm d}$. During training, given an input image $x$, it feeds the randomly masked visual patches into an image encoder and the MAE decoder learns to reconstruct them via $\mathcal{L}_{IR}$. The correlated concepts are also retrieved from a vocabulary for each image and then concatenated with the text $y$ as inputs of the text encoder. The concept-conditioned cross-modal decoder is fed with image features, concept-conditioned text features and text embedding, and optimized via $\mathcal{L}_{LM}$. The noise-adaptive image-text contrastive loss $\mathcal{L}_{NITC}$ is adopted to learn cross-modal alignment by considering varying noise probabilities. Note that the concept-conditioned cross-modal decoder does not utilize image tokens as input for $\mathcal{L}_{NITC}$ to avoid information leakage while does for $\mathcal{L}_{LM}$.
		Omit the index $i$ here.}
		\label{fig:architecture}
\end{figure*}

\noindent\textbf{Noisy Data Learning} has been a long-standing research area to cope with the noise in training data, practically all of which are applied to the classification task. Existing studies~\cite{song2020learning} frequently use robust architecture design, regularization, loss modification, or sample selection strategies to limit the detrimental impact of noisy labels. Here we discuss the last three techniques, which are the most relevant to our model. First, the regularization enforces the networks to over-fit less to false-labeled examples explicitly or implicitly, \emph{e.g.}, label smoothing~\cite{pereyra2017regularizing,lukasik2020does} avoids over-fitting by preventing the networks from assigning full probabilities to noisy data samples. Second, the loss modification adjusts the contribution of clean and noisy samples to the loss~\cite{reed2014training,zheng2020error}. Third, sample selection methods concentrate on choosing clean samples from noisy ones. For example, \citet{arpit2017closer} demonstrates the memorization effect of networks that always prefer to learn simple samples before fitting noisy data. Motivated by the memorization effect, \citet{arazo2019unsupervised} adopts a two-component Gaussian Mixture Model (GMM) to fit per-sample loss and treats the samples with minor loss as clean samples.
To transfer the above noisy label learning technique from the classification problem to the cross-matching problem, \citet{huang2021learning} proposes noisy correspondence learning. 
\citet{amrani2021noise} use density of similarity to estimate the noise probability. \citet{thomas2022emphasizing} apply semantic neighborhood discrepancy and diversity to capture the degree of abstractness of an image-text pair.
%It points out that practically all existing cross-modal learning works learning based on an implicit data assumption, \textit{i.e.}, the training data are correctly aligned, while it is inevitable to collect mismatched or even wrong pairs crawled from the web. 
Different from them, NLIP introduces a new noise-adaptive image-text contrastive loss that harmonizes the cross-modal alignment by considering the varying noise probabilities of different pairs and also rectifies the noisy samples via a concept-guided captioner. NLIP would be one of the early attempts that provide effective and efficient schemes within a large-scale image-text pre-training framework. It can be coupled with any VLP models to improve their robustness. 

\section{Method}

We proposed \textbf{N}oise-robust \textbf{L}anguage-\textbf{I}mage \textbf{P}re-training framework~(\textbf{NLIP}), a new VLP framework to learn from noisy image-text pairs. In this section, we first introduce the overall model architecture of NLIP~(Sec.~\ref{sec:architecture}).
Then we present the model details in two noisy learning schemes respectively, including the \emph{noise-harmonization} scheme to harmonize the cross-modal alignment with noise-adaptive
regularization~(Sec.~\ref{sec:noise-adaptive}) and the \emph{noise-completion} scheme to enrich the missing object information of text~(Sec.~\ref{sec:captioner}).

\noindent\textbf{Basic Notations.} We use $D = \{X,Y\}$ to denote the image-text dataset with the images $X=\{x_i\}_{i=1}^N$ and texts $Y=\{y_i\}_{i=1}^N$, where $N$ denotes the total number of image-text pairs of the dataset.
For vision modality, $\mathcal{V}_{\rm e}$ and $\mathcal{V}_{\rm d}$ denote vision encoder and vision decoder respectively. For language modality, $\mathcal{T}_{\rm e}$ denotes the text encoder. We denote the concept-conditioned cross-modal decoder by $\mathcal{C}_{\rm d}$.
\subsection{Overall Architecture}
\label{sec:architecture}
Fig.~\ref{fig:architecture} illustrates an overview of NLIP architecture for learning the high-quality cross-modal feature alignment. 
NLIP contains a visual encoder-decoder inspired by MAE~\cite{he2021masked} for reducing the computation cost and maintaining the high quality of visual feature representation, a text encoder encoding the texts enriched by extra auxiliary visual concepts and a concept-conditioned cross-modal decoder learning to synthesize semantic-consistent captions to complete noisy ones.
For visual modality, we use Vision Transformer(ViT)~\cite{dosovitskiy2020image} that takes the concatenation of an extra [CLS] token embedding and linearly projected image patches as input and output the [CLS] token to represent the global image feature. Specifically, we randomly mask the patches and skip the mask token to reduce the computation cost. To enhance visual feature representation via self-supervised regularization, an MAE decoder is adopted to restore masked patches by Image Reconstruction~(\textbf{IR}) loss $\mathcal{L}_{IR}$: 
\begin{equation}
\mathcal{L}_{IR} = \sum_{i=1}^N(\frac{\mathcal{V}_e(x'_i)}{\lVert \mathcal{V}_e(x'_i)\rVert} - \frac{x_i}{\lVert x_i\rVert})^2.
\end{equation}
where $\lVert \cdot \rVert$ denotes the normalization, and $x'$ represents masked patches.
As for the language modality, we exploit an encoder-decoder structure to obtain the generation capability and synthesize enriched captions. 
We first retrieve the visual concepts (\emph{i.e.}, names of existing objects) for each input image from a large corpus via a pre-trained model. The visual concepts concatenated with corresponding input texts are encoded by text encoder.
Then a concept-conditioned cross-modal decoder is trained with the Language Modeling~(\textbf{LM}) loss $\mathcal{L}_{LM}$ to generate a more detailed caption for each image guided by the visual concepts.
For the cross-modal alignment, the Noise-adaptive Image-Text Contrastive~(\textbf{NITC}) loss $\mathcal{L}_{NITC}$ is conducted to not only encourage the positive pair representations to get closer contrast to the negative pairs but also introduce the noise-adaptive label smoothing as an instance-aware regularization for avoiding severe bias to the noisy data. 
Therefore, the overall loss can be written as:
\begin{align}
\mathcal{L}&= \mathcal{L}_{IR}+\alpha\cdot\mathcal{L}_{LM}+\beta\cdot\mathcal{L}_{NITC}.
\end{align} 
where $\alpha$ and $\beta$ denote the weighting factors.

\subsection{Noise Harmonization}
\label{sec:noise-adaptive}
To avoid over-fitting to the noisy image-text pairs, NLIP introduces the noise harmonization scheme by learning to harmonize the cross-modal alignments and adopts noise-adaptive regularization for each pair based on the estimated noisy probability.

\noindent\textbf{Preliminaries. }To align between two different modalities, 
current vision-language pre-training models~\cite{radford2021learning} adopt the Image-Text Contrastive~(ITC) loss, to encourage positive image-text pairs $\{x_i, y_j\}_{i=j}$ aligned in the same feature space while in contrast to the negative pairs $\{x_i, y_j\}_{i \neq j}$.
The normalized features from the image encoder and text encoder are denoted as $\mathcal{V}_e(x_i)$ and $\mathcal{T}_e(y_i)$. We first calculate the per-sample image-to-text similarity $s^y \in \mathbb{R} ^ {B \times B}$ and text-to-image similarity $s^x \in \mathbb{R} ^ {B \times B}$ in a batch as:
\begin{align}
    s^y_{i,j} = s^x_{i,j} &= \mathcal{V}_e(x_i)^\top\mathcal{T}_e(y_j).
\end{align}
where $B$ denotes the batch size.
Then the Image-Text Contrastive loss $\mathcal{L}_{ITC}$ can be written as the average of image-to-text and text-to-image contrastive loss:
\begin{align}
\mathcal{L}_{ITC} &= \frac{1}{2B}\sum^{B}_{i=1}(\mathcal{L}_i^x + \mathcal{L}_i^y), \\
\mathcal{L}_i^x=\mathcal{L}_i^x(x_i,\{y_j\}_{j=1}^B)&=-\log \frac{\exp (s_{i,i}^x)}{ \sum_{j} \exp (s_{i,j}^x)}, \\
\mathcal{L}_i^y=\mathcal{L}_i^y(y_i,\{x_j\}_{j=1}^B)&=-\log \frac{\exp (s_{i,i}^y)}{ \sum_{j} \exp (s_{i,j}^y)}. 
\end{align}
However, existing ITC loss forces models to align the feature of each image-text pair without considering the situation that many of them are noisy. Directly pre-training with these samples may degrade the model performance. 

\noindent\textbf{Noise-adaptive Image-Text Contrastive Loss.}
We further propose a Noise-adaptive Image-Text Contrastive~(NITC) loss $\mathcal{L}_{NITC}$ to harmonize the cross-modal alignments with varying degrees according to its noisy probability. We first calculate the noisy probability of each image-text pair, which indicates the image and text in this pair are not semantically matched, according to the memorization effect~\cite{arpit2017closer,zhang2021understanding}. Specifically, the cross-modal transformer tends to fit the easy (\emph{i.e.}, clean) samples first and then the noisy ones.
Therefore, we adopt a two-component Gaussian Mixture Model~(GMM)~\cite{permuter2006study} to fit the per-sample ITC loss. Specifically, we consider the probability predicted by the higher mean component as noisy probability $\varepsilon_i$ of $i$-th image-text pair, inspired by~\cite{huang2021learning,arazo2019unsupervised}:
\begin{align}
\label{eq:gmm} p(\mathcal{L}_{ITC}(x_i,y_i)|\theta)
% \varepsilon_i 
= \sum^2_{m=1}\gamma_m\phi(\mathcal{L}_{ITC}(x_i,y_i)|m), \\ 
\varepsilon_i 
% = p(m_l|\mathcal{L}_{ITC}(x_i,y_i))
=p(\mu_h)p(\mathcal{L}_{ITC}(x_i,y_i)|\mu_h)/p(\mathcal{L}_{ITC}(x_i,y_i)).
\end{align}
where $\gamma_m$ denotes the mixture coefficient, $\phi(\cdot|m)$ is the probability density of the $m$-th GMM component, $\theta$ represents the parameters of GMM, and $\mu_h$ denotes the component with a higher mean.

Then we directly regularize the ground-truth alignment label with various degrees considering its noisy probability $\varepsilon_i$.
Lower regularization is adopted for the clean samples~(\emph{i.e.}, with low $\varepsilon_i$) to learn the alignment, while the higher regularization is adopted for noisy samples~(\emph{i.e.}, with high $\varepsilon_i$) to avoid over-fitting the noise.
In detail, inspired by the label-smoothing~\cite{szegedy2016rethinking}, we regularize the ground-truth image-to-text and text-to-image alignment label with different smoothing rates $W=\{w_i\}_{i=1}^N$, which is linearly associated with the noisy probability of each sample \{$w_i = \lambda \varepsilon_i,\ w_i \in [0,\lambda]$\}. $\lambda$ denotes the hyper-parameter to control the range of smooth rate.
Then the Noise-adaptive Image-Text Contrastive loss $\mathcal{L}_{NITC}$ is defined as:
% {\fontsize{6.5pt}{\baselineskip}\selectfont
% \begin{align}
% \mathcal{L}_{NITC} &= \frac{1}{2B}\sum^{B}_{i=1}(\hat{\mathcal{L}}_i^x + \hat{\mathcal{L}}_i^y), \\
% \hat{\mathcal{L}}_i^x(x_i,\{y_j\}_{j=1}^B)&=-\log \frac{(1 - w_i) \exp (s_{i,i}^x)}{(1 - w_i) \exp{(s_{i,i}^x)}\!+\!\frac{w_i}{B-1} \sum\limits_{i\neq j} \exp (s_{i,j}^x)},\\
% \hat{\mathcal{L}}_i^y(y_i,\{x_j\}_{j=1}^B)&=-\log \frac{(1 - w_i) \exp (s_{i,i}^y)}{(1 - w_i) \exp{(s_{i,i}^y)}\!+\!\frac{w_i}{B-1} \sum\limits_{i\neq j} \exp (s_{i,j}^y)}.
% \end{align}}
{\fontsize{8pt}{\baselineskip}\selectfont
\begin{align}
\mathcal{L}_{NITC} &= \frac{1}{2B}\sum^{B}_{i=1}(\hat{\mathcal{L}}_i^x + \hat{\mathcal{L}}_i^y), \\
\hat{\mathcal{L}}_i^x&=-\log \frac{(1 - w_i) \exp (s_{i,i}^x)}{(1 - w_i) \exp{(s_{i,i}^x)}\!+\!\frac{w_i}{B-1} \sum\limits_{i\neq j} \exp (s_{i,j}^x)},\\
\hat{\mathcal{L}}_i^y&=-\log \frac{(1 - w_i) \exp (s_{i,i}^y)}{(1 - w_i) \exp{(s_{i,i}^y)}\!+\!\frac{w_i}{B-1} \sum\limits_{i\neq j} \exp (s_{i,j}^y)}.
\end{align}}\subsection{Noise Completion}
\label{sec:captioner}
Apart from adopting the above instance-ware regularization on the noisy pairs, NLIP also introduces the noise completion scheme to enrich the missing object information of text since the captions from the web are naturally incomplete. 
Especially, NLIP injects a concept-conditioned cross-modal decoder to obtain semantic-consistent synthetic captions to complete noisy ones, which uses the retrieved visual concepts (\emph{i.e.}, names of existing objects) for the corresponding image to guide captioning generation.

\noindent\textbf{Visual Concept.}
\label{sec:cues}
Although the image-text data can be easily crawled from the web, the texts usually contain much noise, including missing details of the image and carrying unrelated contents to the image~\cite{li2022blip}. 
To better address the problem of image-text misalignment, we introduce the visual concepts $q_{\rm v}$ as auxiliary inputs to provide the prior information of existing objects for each image.
We first construct a large visual concept vocabulary $Q$ via parsing the various concept nouns from the web-collected corpus.
Then we retrieve the words of top-k similarity with image $x_i$ as visual concepts $q_i \in Q$ based on a pre-trained VLM for that image. 
The similarity $sim(x_i,Q)$ between the input image $x_i$ and the nouns in $Q$ is calculated by 
\begin{align}
sim(x_i,Q) &= \langle\mathcal{V}_{\rm e}(x)\cdot \mathcal{T}_{\rm e}([{\rm{p}},Q])\rangle.
\end{align}
where ${\rm{p}}$ denotes the pre-defined text prompt that is aggregated with the visual concepts to narrow down the gap with natural language~\cite{radford2021learning}.
Based on the retrieved visual concepts $q_i$, NLIP uses an additional concept-conditioned cross-modal decoder~(shown in Fig.~\ref{fig:architecture}) to synthesize new texts $Y'$ to replace the original texts $Y$ in noisy image-text pairs. 
Specifically, the cross-modal decoder is optimized by recovering the masked texts $y^m$ with an autoregressive (\textit{i.e.}, language modeling) loss: 
{\fontsize{7.8pt}{\baselineskip}\selectfont
\begin{align}
\mathcal{L}_{LM}&= -\mathbb{E}_{(x,y)\sim D}\log p(y_t|\mathcal{C}_{\rm d}(y_{\tau<t},[\mathcal{V}_{\rm e}(x),\mathcal{T}_{\rm e}([{\rm{p}},q,y^m])])).
\end{align}
}
where $[\cdot]$ denotes the concatenation operation and $t$ denotes the word index of text $y$. Note that we omit index $i$ here.

\begin{figure}[t!]
        \centering
        \includegraphics[width=\linewidth]{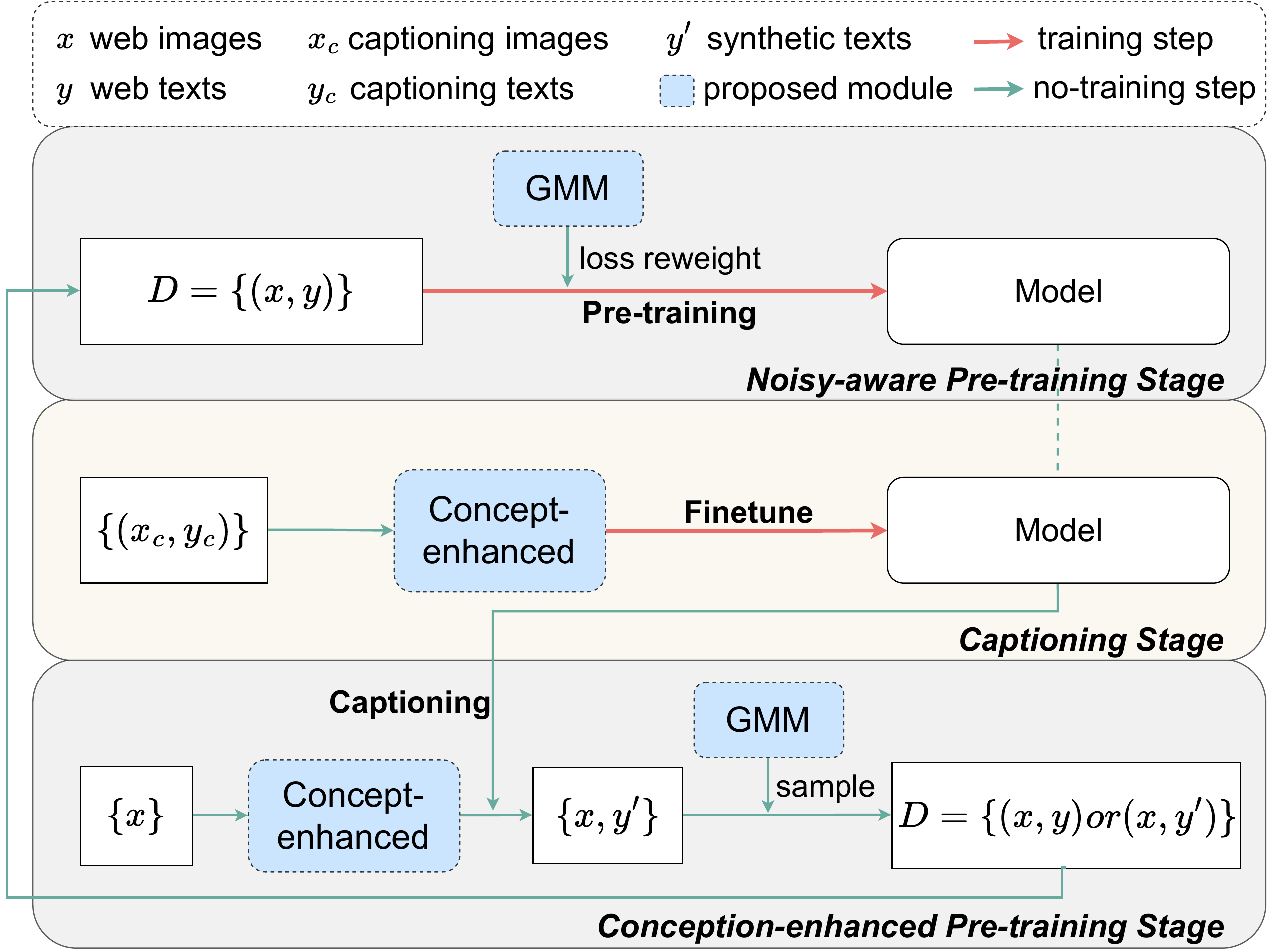}
		\caption{Illustration of NLIP procedure. The whole pre-training contains three stages: \emph{noisy-aware pre-training}, \emph{captioning} and \emph{conception-enhanced pre-training}. At \emph{noisy-aware pre-training} stage, we adopt the noisy-adaptive regularization to pre-train NLIP. At \emph{captioning} stage, we use captioning data to train concept-conditioned cross-modal decoder and generate synthetic captions for web images. At \emph{conception-enhanced pre-training} stage, we select training captions by noisy probabilities and fine-tune NLIP.}
		\label{fig:procedure_visual concept}
\end{figure}

\subsection{Pre-training procedure}
\label{sec:pre-training procedure}
As shown in Fig.~\ref{fig:procedure_visual concept}, we divide the whole pre-training paradigm of NLIP into three steps: \emph{noisy-aware pre-training}, \emph{captioning} and \emph{conception-enhanced pre-training}. At \emph{noisy-aware pre-training} stage, we first warm up the NLIP architecture with $E_e$ epochs under the supervision of $\mathcal{L}_{IR}$, $\mathcal{L}_{LM}$ and $\mathcal{L}_{ITC}$. 
Then we estimate the noisy probability $\varepsilon_i$ of the $i$-th image-text pair based on the $\mathcal{L}_{ITC}$ and adopt the noisy-adaptive regularization by replacing the $\mathcal{L}_{ITC}$ with $\mathcal{L}_{NITC}$ in the following $E_t$ epochs.
At \emph{captioning} stage, to obtain better generation ability, we further fine-tune the captioner, which includes the image encoder $\mathcal{V}_{\rm e}$, text encoder $\mathcal{T}_{\rm e}$ and cross-modal decoder $\mathcal{C}_{\rm d}$, on captioning dataset COCO Captions~\cite{lin2014microsoft} and generates new texts $Y'=\{y'_i\}_{i=1}^N$ for each image-text pair. Finally, at \emph{conception-enhanced pre-training} stage, we fine-tune NLIP with $E_f$ epochs with the revised image-text pairs $D'$, where each text $y_i$ of the $i$-th pair in original dataset $D$ is replaced by the synthetic text $y'_i$ randomly with sampling rate same as the noisy probability $\varepsilon_i$.

\begin{table*}[t!]
\caption{Top-1 accuracy(\%) of zero-shot image classification and linear probing image classification tasks on 12 datasets when pre-training on YFCC26M. 
\label{tab:zeroshot-classification-table1}}
\centering
\resizebox{0.98\textwidth}{!}{
\begin{tabular}{l|c|cccc cccc cccc |c}
& \rotatebox{75}{Backbone}
&\rotatebox{75}{CIFAR10}~~ &
\rotatebox{75}{CIFAR100}~~ &
\rotatebox{75}{Caltech101}~~ &
\rotatebox{75}{StanfordCars}~~ &
\rotatebox{75}{Flowers102}~~ &
\rotatebox{75}{Food101}~~ &
\rotatebox{75}{SUN397}~~ &
\rotatebox{75}{DTD}~ &
\rotatebox{75}{Aircrafts}~~ &
\rotatebox{75}{OxfordPets}~~ &
\rotatebox{75}{EuroSAT}~~ & 
\rotatebox{75}{\textbf{ImageNet}}~~ &
\rotatebox{75}{\textbf{Average}}~~ \\
\noalign{\smallskip}
\hline
\noalign{\smallskip}
\multicolumn{15}{c}{\textbf{Zero-Shot Image Classification}}\\
\noalign{\smallskip}
\hline
\noalign{\smallskip}
CLIP & \multicolumn{1}{c|}{\multirow{3}{*}{ViT-B/32}}& 
74.8&
44.1&
64.5&
3.7&
51.4&
45.1&
43.7&
14.5&
4.3&
22.9&
23.0&
34.8&
35.6\\
FILIP &
& 
\textbf{83.6} & 
\textbf{51.7} & 
73.6 & 
\textbf{7.8} & 
\textbf{60.5} & 
\textbf{55.9} & 
47.9 & 
18.8 & 
8.0 & 
29.9 & 
29.5 & 
41.4 & 
42.4 \\
NLIP & & 74.0 &  47.4 & \textbf{75.1} &  6.8 &  58.9 & 53.8 &  \textbf{55.4} &  \textbf{32.3} &  \textbf{8.9} & \textbf{36.8} &  \textbf{35.4} &  \textbf{42.4}& \textbf{43.9} \\
\noalign{\smallskip}
\hline
\noalign{\smallskip}
 CLIP & \multicolumn{1}{c|}{\multirow{3}{*}{ViT-B/16}}  & 75.3
 & 42.4  & 69.5 & 3.9 & 54.8
 & 51.1
 & 46.6 & 18.6 & 3.9 & 21.7
 & 20.5 & 39.2 & 37.3  \\
 FILIP &  & \textbf{83.8} &
\textbf{51.2} &
76.1 &
\textbf{8.9} &
\textbf{62.8 }&
\textbf{63.5} &
52.5 &
21.8 &
\textbf{10.2} &
36.7 &
24.9 &
46.7 &
44.9 \\
NLIP & & 81.9 & 47.5 & \textbf{79.5} & 7.8 & 54.0 & 59.2 & \textbf{58.7} & \textbf{32.9} & 7.5 & \textbf{39.2} & \textbf{33.9} & \textbf{47.4} & $\textbf{45.9}$ \\
\noalign{\smallskip}
\hline
\noalign{\smallskip}
\multicolumn{15}{c}{\textbf{Linear Probing Image Classification}}\\
\noalign{\smallskip}
\hline
\noalign{\smallskip}
CLIP & \multicolumn{1}{c|}{\multirow{3}{*}{ViT-B/32}} &
90.4 &
69.7&
84.7&
23.8&
91.5&
70.7&
66.3&
66.1 &
32.7&
61.0&
96.0&
60.3&
67.8 \\
FILIP &
& 90.5 & 69.5 & 88.2 & 30.0 & 90.9 & 69.2 & 67.6 & 66.0 & 31.3 & 56.0 & 93.4 & 58.8 & 67.6 \\
NLIP &  & \textbf{90.9}& 
\textbf{73.4}& 
\textbf{89.2}& 
\textbf{34.1}& 
\textbf{95.6}& 
\textbf{76.9}& 
\textbf{71.9}& 
\textbf{71.3} &
\textbf{39.8}& 
\textbf{62.5}& 
\textbf{96.8}& 
\textbf{67.1}& 
\textbf{72.5} \\
\noalign{\smallskip}
\hline
\noalign{\smallskip}
 CLIP & \multicolumn{1}{c|}{\multirow{3}{*}{ViT-B/16}} &
 90.5 & 71.1 & 86.6 & 29.4 & 92.8 & 78.4 & 67.7 & 66.2 & 37.2 & 66.0 & 94.3 & 65.0 & 70.4  \\
 FILIP & & 
 90.6& 
67.4& 
88.6& 
32.8& 
93.7& 
71.8& 
69.8& 
68.5& 
35.7& 
59.4& 
93.7& 
62.3& 
69.5 \\
NLIP & & \textbf{92.8} & \textbf{74.2} & \textbf{90.4} & \textbf{41.2} & \textbf{97.5} & \textbf{85.0} & \textbf{75.9} & \textbf{74.3} & \textbf{43.4} & \textbf{79.2} & \textbf{96.8} & \textbf{71.8} & \textbf{76.9} \\
\noalign{\smallskip}
\hline
\end{tabular}
}
\end{table*}

\section{Experiments}
\label{sec:exps}
\subsection{Experimental Settings}
\label{sec:experiment_details}
\textbf{Model Architecture.}
We adopt the ViT-B/16 and ViT-B/32 as our visual encoder architecture.
Unless specified, NLIP uses ViT-B/16 as the visual encoder. The text encoder and concept-conditioned cross-modal decoder are initialized from $\text{BART}_{\text{base}}$~\cite{lewis2020bart} and the MAE decoder only has 4 transformer blocks with 64-d head.

\noindent \textbf{Training Details.}
We pre-train our NLIP on 32 Nvidia V100 for 50 epochs with 6144 batch size. 
LAMB~\cite{2019Large} optimizer is adopted with a weight decay of 0.05. 
The base learning rate is set to 0.003 and the scaling rule keeps the same with~\citet{yao2021filip}. 
The learning rate is linearly warmed up in the first five epochs and then gets decayed by the cosine learning rate schedule~\cite{2016SGDR}. 
We pre-train NLIP on a 26M subset of YFCC100M named YFCC26M, and the filtering rules follow FILIP~\cite{yao2021filip}. 
% The visual concept cues into the text encoder with 0.5 probability during pre-training. 
During the pre-training, the images are randomly cropped between 50\% and 100\% of the original size and then resized to 224 $\times$ 224 resolution. The visual encoder applies 50\% masking ratio.
When conducting downstream tasks (\emph{e.g.}, image captioning), the image resolution is resized to 384 $\times$ 384 and we don't mask any image patches.
The training epochs $E_e$, $E_t$ and $E_f$ in different stages are set as 5, 45 and 20, respectively. 
The weighting factor $\alpha$ and $\beta$ are both 1 and $\lambda$ in $\mathcal{L}_{NITC}$ is 0.5. 
During \emph{captioning} stage, following BLIP~\cite{li2022blip}, we fine-tune NLIP on COCO~\cite{lin2014microsoft}'s Karpathy train split~\cite{karpathy2015deep}
to generate high-quality captions. 
Note that the COCO Captions contains only 113K images and 567K human-annotated caption while YFCC26M contains 230x more data than COCO Captions.
We discuss the effect of fine-tuning captioner with COCO Captions in Appendix.

\noindent\textbf{Visual Concept Vocabulary.}
The visual concept vocabulary $Q$ is built by parsing the nouns from the collected text corpus via spaCy toolkit and filtering nouns that appear less than 5 times. 
The source corpus includes YFCC100M~\cite{thomee2016yfcc100m}, OpenWebText~\cite{gokaslan2019openwebtext}, WordNet of NLTK~(Natural Language Toolkit)~\cite{loper2002nltk} and the most-frequent n-gram collected from web. 

After collecting, the visual concept vocabulary $Q$ contains about 151k unique nouns.
We use a pre-trained $\text{FILIP}_\text{large}$~\cite{yao2021filip} to retrieve visual concepts for each image. Unless specified, NLIP uses $\text{FILIP}_\text{large}$ to retrieve visual concepts.
More ablation studies about the effect of utilizing 
different pre-trained VLMs (e.g. YFCC26m-pretrained CLIP-ViT-L/16 and CLIP-ViT-L/14)
and the effect of different visual concept vocabularies
are shown in Sec. \ref{sec:ablation_study} and Appendix.
\subsection{Image Classification}
We evaluate our proposed NLIP on the zero-shot image classification and linear probing image classification tasks on 12 downstream classification datasets as in Table \ref{tab:zeroshot-classification-table1}, demonstrating the superior zero-shot transfer capability.
These 12 classification datasets consist of CIFAR10~\cite{krizhevsky2009learning}, CIFAR100~\cite{krizhevsky2009learning}, Caltech101~\cite{fei2006one}, StanfordCars~\cite{krause20133d}, Flowers102~\cite{nilsback2008automated}, Food101~\cite{bossard2014food}, SUN397~\cite{xiao2010sun}, DTD~\cite{cimpoi2014describing}, Aircrafts~\cite{maji2013fine}, OxfordPets~\cite{parkhi2012cats}, EuroSAT~\cite{helber2019eurosat}, ImageNet~\cite{russakovsky2015imagenet}, covering a wide range of domains.
Note that the linear probing task only trains a random initialized linear classifier with a frozen image encoder on the downstream datasets.
We compare with other vision-language pre-training methods, including FILIP with  the token reduction layer~\cite{yao2021filip,gu2022wukong} and CLIP~\cite{radford2021learning} under the same  dataset (\emph{i.e.}, YFCC26M) and the same evaluation settings in~\cite{radford2021learning}. 
For fair comparison, we pre-train CLIP with the same augmentation strategies as ours on YFCC26M. We ensemble all prompts by averaging the text embeddings for each class across the prompt templates as in~\cite{radford2021learning} for all models. More results in CLIP benchmark~\cite{cui2022democratizing} are listed in Appendix.

\noindent\textbf{Zero-Shot Image Classification.} Experimental results show that NLIP largely outperforms the corresponding baseline CLIP in terms of average top-1 accuracy over 12 datasets and achieves an improvement of 8.6\%. In particular, NLIP surpasses CLIP on ImageNet over 8.2\%. Besides, NLIP also obtains substantial performance gains in most individual datasets with images in different domains,
demonstrating the effectiveness of proposed noise-harmonization and noise completion schemes.
Compare to FILIP which learns the finer-grained alignment between image and text, NLIP with global image-text alignment achieves 1.0\% average improvement over 12 datasets. 

\noindent\textbf{Linear Probing Image Classification.} Table \ref{tab:zeroshot-classification-table1} demonstrates that NLIP achieves 76.9\% on average top-1 accuracy over 12 downstream tasks, which surpasses FILIP and CLIP by 7.4\% and 6.5\%, respectively. NLIP with ViT-B/32 also outperforms FILIP and CLIP about 4.9\% and 4.7\%. The linear probing experiments demonstrate the robustness representation learned by NLIP.

\subsection{Image-Text Retrieval}
We evaluate NLIP on both zero-shot image-to-text retrieval (TR) and zero-shot text-to-image retrieval (IR) tasks on Flickr30K~\cite{dataset_flickr30k}. Then we also compare NLIP against the existing vision-language pre-training methods, including Unicoder-VL~\cite{li2020unicoder}, ImageBERT~\cite{qi2020imagebert}, UNITER~\cite{chen2020uniter}. 
These models are single-stream and employ an additional object detector to extract region features while NLIP only employs visual patch features for simplicity. 

As shown in Table \ref{tab:zero-shot-retrieval-table}, NLIP achieves substantial improvement compared with CLIP pre-trained in YFCC26M. In image-to-text retrieval, NLIP outperforms CLIP 28.7\% in R@1.
In text-to-image retrieval, NLIP is 26.6\% higher than CLIP on R@1 and 7.1\% higher than CLIP* fine-tuned on MSCOCO dataset.
NLIP also achieves 1.9\% improvement than UNITER in R@1. As shown in Table~\ref{tab:full-ablation-study-table}, when only using YFCC26M-pretrained CLIP to retrieve visual concepts, our NLIP still beats CLIP and CLIP* over 23.2\% and 3.6\% on zero-shot image-to-text retrieval task, which demonstrates the superiority of the noise-robust learning in NLIP under the exact same pre-training data.

\subsection{Image Captioning}
We further evaluate the pre-trained NLIP on downstream image captioning task, which aims at generating the description of an image in natural language, on COCO Caption~\cite{lin2014microsoft} dataset. 
We evaluate different methods on standard metrics for the captioning task, including BLEU~\cite{papineni2002bleu}, CIDEr~\cite{vedantam2015cider}. 
For fair comparison with other models, we follow BLIP~\cite{li2022blip} to initialize the visual encoder of NLIP from an ImageNet pre-trained ViT-B/16. 

As shown in Table \ref{tab:caption}, NLIP achieves 40.3 in BLEU@4 and 135.2 in CIDEr, outperforming BLIP~\cite{li2022blip} by 1.9 in CIDEr. Note that BLIP is pre-trained with 5x more image-text pairs(129M v.s. 26M). NLIP with train-from-scratch image encoder still outperforms BLIP, according to the third row of Table. \ref{tab:full-ablation-study-table}. NLIP also beats other methods~(\emph{e.g.}, SimVLM) pre-trained on large-scale datasets. Particularly, VinVL~\cite{zhang2021vinvl} requires an object detector pre-trained on 2.5M images with high resolution (800×1333) and full human-annotated bounding boxes.

\begin{table}
\caption{Results of zero-shot image-to-text and text-to-image retrieval on Flickr30K. * means the model fine-tuned on MSCOCO dataset.
}

\label{tab:zero-shot-retrieval-table}
\centering
\resizebox{\linewidth}{!}{{
\setlength{\tabcolsep}{0.5em}
    {\renewcommand{\arraystretch}{1.0}
\begin{tabular}{lcccccccccccc}
\hline
\noalign{\smallskip} 
            & \multicolumn{3}{c}{image-to-text} & \multicolumn{3}{c}{text-to-image}\\
            & R@1           & R@5           & R@10         & R@1           & R@5           & R@10        \\
\noalign{\smallskip}   
\hline
\noalign{\smallskip}   
Unicoder-VL~\cite{li2020unicoder} & 64.3 & 85.8 & 92.3 & 48.4 &  76.0 & 85.2 
\\
ImageBERT~\cite{qi2020imagebert}  & 70.7 & 90.2 & 94.0 & 54.3 &  79.6  & 87.5
\\
UNITER~\cite{chen2020uniter} & 80.7 & 95.7 & 98.0 & 66.2 &  88.4 &  92.9 \\
% {\color{gray}UNITER~\cite{chen2020uniter}} &{\color{gray} 80.7} & {\color{gray}95.7} & {\color{gray} 98.0} & {\color{gray} 66.2} & {\color{gray} 88.4} & {\color{gray} 92.9} \\
\noalign{\smallskip}   
\hline
\noalign{\smallskip}
CLIP(ViT-B/32) & 46.4 & 75.4 & 84.1 & 29.8 & 56.1 & 67.8 \\
FILIP(ViT-B/32) & 56.6 & 82.7 & 90.0 & 39.5 & 66.7 & 75.8 \\
\rowcolor{mypink} \textbf{NLIP(ViT-B/32)} & \textbf{77.2} & \textbf{94.8} & \textbf{97.7} & \textbf{56.6}  & \textbf{83.2}  & \textbf{89.8} \\
\noalign{\smallskip}   
\hline
\noalign{\smallskip}
CLIP(ViT-B/16)  & 53.9 & 81.0 & 90.1 & 34.6 & 62.6 & 73.6 \\
CLIP(ViT-B/16)* & 73.5 & 92.6 & 96.2 & 54.1 & 81.9 & 89.8 \\
FILIP(ViT-B/16) & 66.5 & 88.4 & 93.9 & 47.1 & 74.4 & 82.5 \\
\rowcolor{mypink} \textbf{NLIP(ViT-B/16)} & \textbf{82.6} & \textbf{96.6} & \textbf{98.3} & \textbf{61.2}  & \textbf{85.7}  & \textbf{91.7} \\
\hline
\end{tabular}}}}
\end{table}

\begin{table}[t!]
\caption{Comparison with SoTA image captioning methods on COCO captioning benchmark. NLIP achieves the best performance even using a small-scale pre-training dataset.}
\label{tab:caption}
\centering
\resizebox{0.98\linewidth}{!}{{
\setlength{\tabcolsep}{0.5em}
{\renewcommand{\arraystretch}{1.0}
\label{ablation-yfcc-table}
\begin{tabular}{l|c|cccc}
\hline
\noalign{\smallskip} 
\multirow{2}{*}{Model} & $\#$ Pre-train & \multicolumn{2}{c} { MSCOCO } \\ 
   & Images & BLEU@4 & CIDEr\\  % & B@4 & C
\noalign{\smallskip}
\hline
\noalign{\smallskip} 
Encoder-Decoder~\cite{changpinyo2021cc12m} & 15M  & - & 110.9  \\
BUTD~\cite{anderson2018bottom} & 1.7M & 36.4 & 120.1 \\
VinVL\cite{zhang2021vinvl} & 5.7M  & 38.2 & 129.3  \\
VLP~\cite{zhou2020unified} & 3M & 39.5 & 129.8\\
AoANet~\cite{huang2019attention} & 1.7M & 38.9  & 129.8\\ 
$\text{UNIMO}_{\text{base}}$~\cite{li2020unimo} & 11.3M & 38.8   & 124.4 \\
$\text{SimVLM}_{\text{base}}$~\cite{wang2021simvlm} & 1.8B & 39.0  & 134.8 \\
BLIP~\cite{li2022blip} & 129M & 39.7 & 133.3 \\ 
\hline
\rowcolor{mypink} NLIP & 26M & \textbf{40.3} & \textbf{135.2}  \\
\hline
\end{tabular}}}}
\end{table}

\subsection{Ablation Studies}
\label{sec:ablation_study}
\noindent\textbf{Effect of Noise Harmonization.}
Table \ref{tab:full-ablation-study-table} ablates the effectiveness of our noise harmonization. By comparing with the last two rows, we can find that NLIP gains 1.2\% and 2.5\% improvement in image-to-text retrieval and text-to-image retrieval with noise harmonization, respectively, verifying that pre-training with NITC loss helps the model avoid over-fitting on the mismatched image-text pairs.

\noindent\textbf{Effect of Noise Completion.} 
Table~\ref{tab:full-ablation-study-table} shows that NLIP with the noise completion scheme can boost performance on all downstream tasks.
We can observe the noise completion scheme helps boost the image caption task by over 1.6\% on CIDEr and the text retrieval task by 10.4\% on R@1. Besides, without the condition of visual concepts in noise completion, NLIP will drop 0.7\% accuracy on zero-shot ImageNet classification and 1.1\% R@1 on image retrieval.
Incorporating visual concepts into the cross-modal decoder further help enrich the synthetic caption with more information of existing objects and boost the performance in all downstream tasks, as shown in Table~\ref{tab:full-ablation-study-table}. We illustrate some examples via noise completion in Appendix.

\begin{table}
\caption{Ablation studies of all components on zero-shot classification, image-text retrieval and image caption. We denote using condition of visual concepts in noise completion as ``VC'', noise completion as ``NC'', and noise harmonization as ``NH''. Note that removing the noise completion scheme degrades the performance severely. $\dagger$ denotes using the YFCC26M-pretrained CLIP to retrieve visual concepts.}
\label{tab:full-ablation-study-table}
\centering
\resizebox{0.98\linewidth}{!}{{
\setlength{\tabcolsep}{0.5em}
    {\renewcommand{\arraystretch}{1.0}
\begin{tabular}{lcccccccccccc}
\hline
\noalign{\smallskip} 
        Dataset  & ImageNet & \multicolumn{2}{c}{COCO} & \multicolumn{4}{c}{Flickr30K} \\
        Task & ZS-CLS   & \multirow{2}{*}{ BLEU }  & \multirow{2}{*}{ CIDEr } &  \multicolumn{2}{c}{image-to-text} & \multicolumn{2}{c}{text-to-image} \\
        Metric    & Top-1   &    &   &   R@1                  & R@10         & R@1                & R@10        \\
\noalign{\smallskip}   
\hline
\noalign{\smallskip}
CLIP & 39.2 &-&-& 53.9 & 90.1 & 34.6 & 73.6 \\
\rowcolor{mypink} \textbf{NLIP$\dagger$} & 43.0 & 39.0 & 130.6 & 77.1 & 98.2 & 63.9& 92.5 \\
\rowcolor{mypink} \textbf{NLIP} & 47.4 & 39.9 & 134.0 & 82.6 & 98.3 & 61.2 & 91.7 \\
~ w/o VC & 46.7 & 39.6 & 132.8 & 82.2	& 98.6 & 60.1 &	91.6  \\
~ w/o NC & 47.0 & 39.6 & 132.4 & 72.2	& 96.1 & 49.6 &	84.2  \\
~ w/o NH & 46.7 & 39.6 & 131.5 & 71.0 & 95.6 & 47.1 & 82.0  \\
\hline
\end{tabular}
}}}
\end{table}

\section{Conclusion}
In this paper, we propose a new vision-language pre-training framework named NLIP to learn from the noisy image-text pairs crawled from the web.
NLIP introduces two schemes, including noise-harmonization and noise-completion, to stabilize the pre-training and efficiently make full use of noisy pairs. 
In noise-harmonization scheme, NLIP adopts noise-adaptive regularization to
harmonize the cross-modal alignments with varying degrees by considering the noise probability of each pair. 
And in noise-completion scheme, NLIP further introduces a concept-conditioned cross-modal decoder to obtain synthetic captions to complete noisy ones.
Retrieved visual concepts are utilized as the auxiliary input for the cross-modal decoder to provide the prior information of existing objects. 
Experiments show that NLIP achieves significant performance gaps on several downstream tasks, including zero-shot classification, image-text retrieval and caption generation tasks. In the future, our NLIP can be easily injected into any cross-modal pre-training models and the proposed noisy-robust learning schemes can be beneficial for more downstream fine-grained tasks such as open-world object detection, segmentation, and image generation. 

\section{Acknowledgments}
This research is supported by the Fundamental Research Funds for the Central Universities, Sun Yat-sen University under Grant No. 22lgqb38. We gratefully acknowledge the support of MindSpore\footnote{\url{https://www.mindspore.cn/}}, CANN~(Compute Architecture for Neural Networks) and Ascend AI Processor used for this research.

{
\small
\bibliography{aaai23}
}

\appendix
\section*{Appendix for NLIP: Noise-robust Language-Image Pre-training}

% \hjh{no reference for fig2?}
% \hjh{may summarize first since you don't give the specific chapter number in MAIN paper}
% \hjh{check that all the Appendix in  MAIN paper are addressed}

\section{More Details about Pre-training}
\label{sec:More-Details-about-Pre-training}

For all three stages, NLIP uses Automatic Mixture Precision(AMP) to accelerate the training and gradient checkpoint to save memory and enlarge the batch size. NLIP does not apply the weight decay regularization on embedding, bias and layer normalization. The maximum context length of the text is 77.

During \emph{noisy-aware pre-training} stage, the input of the text encoder is a masked text. NLIP randomly samples a span of the text and replaces the span with a $\langle$mask$\rangle$ token and the span lengths are drawn from a Poisson distribution ($\lambda$ = 3). The input of the text decoder is a complete text without masking. The language modeling loss at the pre-training stage can be regarded as recovering the masked part. Table \ref{tab:pre-training hyperparams} shows the hyperparameters for pre-training. 

\begin{table}[h!]
    \centering
    \caption{More hyperparameters used for NLIP at \emph{pre-training} stage. Vocabulary means the vocabulary of the text encoder. Temperature is a learnable parameter used in contrastive learning.
    % \hjh{give some details, like including the vocabulary size for xxx, the initial temperature for xxx and some hyperparameters for LAMB optimizer}\hrh{show in table}
    }
    % for common hyperparameters.}
    \label{tab:pre-training hyperparams}
    \centering
     \begin{tabular}{l|c}
        \hline
        \noalign{\smallskip}
        Hyperparameter & Value \\
        \noalign{\smallskip}
        \hline
        \noalign{\smallskip}
        Vocabulary size & $50265$ \\
        Initial temperature & $0.07$ \\
        % LAMB $\beta_{1}$ & $0.9$ \\
        % LAMB $\beta_{2}$ & $0.95$ \\
        % LAMB $\epsilon$  & $10^{-4}$ \\
        LAMB beta1 & $0.9$ \\
        LAMB beta2 & $0.95$ \\
        LAMB epsilon  & $10^{-4}$ \\
        Weight decay  & $5e^{-2}$ \\
\noalign{\smallskip}
\hline
    \end{tabular}
\end{table}

During \emph{captioning}
% \xd{be consistent with main paper}
stage, we use Cross-Entropy Optimization and Self-Critical Sequence Training (SCST) \cite{rennie2017self} both. Note that we don't mask any image patches at this stage. We feed the visual concepts aggregated with a prompt prefix `This photo may describe these objects:' into the text encoder only. To generate new texts for YFCC26M, we parse the nouns from the original texts of YFCC26M and concatenate the nouns with visual concepts retrieved by the pre-trained VLM. Table \ref{tab:captioner-crossentropy-hyperparameters} summarizes the hyperparameters of Cross-Entropy optimization. The hyperparameters of SCST are similar to Cross-Entropy optimization except only training 5 epochs and the batch size is set to 96~(\emph{i.e.}, 4 for each GPU). 

During \emph{conception-enhanced pre-training} stage, we mainly follow the hyperparameters of \emph{noisy-aware pre-training} stage but shrink the base learning rate to 0.0003 and turn off the Noise-adaptive Contrastive Learning. We set the warm up iterations to 4000.

\begin{table}
    \centering
    \caption{Hyperparameters of CrossEntropy Optimization at \emph{captioning} stage.}
    \label{tab:captioner-crossentropy-hyperparameters}
    \centering
     \begin{tabular}{l|c}
        \hline
        \noalign{\smallskip}
        Hyperparameter & Value \\
        \noalign{\smallskip}
        \hline
        \noalign{\smallskip}
        Epoch & $30$ \\
        Warm up iter & $2217$ \\
        Learning rate    & $1e^{-5}$ \\
        Batch size       & $256$ \\
        % AdamW $\beta_{1}$ & $0.9$ \\
        % AdamW $\beta_{2}$ & $0.999$ \\
        % AdamW $\epsilon$  & $1e^{-4}$ \\
        AdamW beta1 & $0.9$ \\
        AdamW beta2 & $0.999$ \\
        AdamW epsilon  & $1e^{-4}$ \\
        Weight decay   & $5e^{-2}$
        \\
\noalign{\smallskip}
\hline
    \end{tabular}
\end{table}

\section{More Comparison}
The performance comparisons on CLIP benchmark\cite{cui2022democratizing} are shown in Table \ref{tab:ZS-CLS_benchmark}.
We follow CLIP benchmark pre-train NLIP with ViT-B/32 backbone on YFCC15M-v2\cite{cui2022democratizing} to make a fair comparison. NLIP outperforms DeCLIP\cite{li2022supervision}, which performs six different image-text contrastive supervisions, bringing on computation and communication.
\begin{table}[h]
\centering
\caption{Performance comparison on CLIP benchmark~\cite{cui2022democratizing}.}
\resizebox{\linewidth}{!}{
\begin{tabular}{l|ccccc}
\hline
\noalign{\smallskip}
{Downstream Task} & CLIP & SLIP & FILIP & DeCLIP & NLIP \\ 
\noalign{\smallskip}
\hline
\noalign{\smallskip} 
ImageNet~(ZS-CLS)    &  32.8 &  34.3  &  39.5 & 43.2 &  \textbf{43.5}\\
\noalign{\smallskip}
\hline
\end{tabular}}
\label{tab:ZS-CLS_benchmark}
\end{table}

\section{Discussion of Visual Concept}
\label{sec:discussion-visual-concept}
\subsection{Visual Concept with Different VLMs}
\paragraph{Retrieving via Large-scale Pre-trained VLMs} 
Table~\ref{tab:large-scale-vlms-retrieve-concepts} shows the performance comparison of using different large-scale pre-trained VLMs to retrieve the visual concepts.
Experimental results show that the $\text{FILIP}_\text{large}$(pre-trained on 340M image-text pairs) achieves the best performance that outperforms CLIP with ViT-L/14(pre-trained on 400M image-text pairs) about 0.4\%. 
We speculate that $\text{FILIP}_\text{large}$ with fine-grained interaction between image and text improves the effectiveness of the retrieved visual concepts.
\paragraph{Retrieving via VLMs Pre-trained on YFCC26M} To avoid using large-scale pre-trained VLMs and verify the effectiveness of retrieving visual concepts, we conduct an experiment on CLIP with ViT-B/16 pre-trained on YFCC26M, denote CLIP for simple, and discuss which visual concept vocabulary performs the best. Table \ref{tab:yfcc26m-pretrained-self-retrieve} shows the performance comparison of utilizing visual concepts from different corpus.
We use CLIP pre-trained on YFCC26M to retrieve the visual concepts and pre-train CLIP with the retrieved visual concepts. Experimental results show that using the visual concepts of YFCC achieves the best performance which obtains a 4.5\% improvement than the model without visual concepts and is slightly better than using the ensemble visual concept vocabulary. We speculate that the visual concepts from the pre-trained dataset can better leverage 
the knowledge of VLM pre-trained on the same dataset.     

\begin{table*}[t!]
\caption{Ablation studies of different VLMs to retrieve visual concepts from the ensemble concept vocabulary. The results are Top-1 accuracy(\%) of zero-shot image classification on 12 datasets. All models are pre-trained on YFCC26M and follow CLIP's architecture with ViT-B/32 backbone, MAE decoder and 50\% masking ratio. Note that CLIP is pre-trained on 400M image-text pairs and FILIP is pre-trained on 340M image-text pairs.
The first line is the result without using visual concepts. Only use single prompt: `a photo of a [classname]'.
}
\label{tab:large-scale-vlms-retrieve-concepts}
\centering
\resizebox{\textwidth}{!}{
\begin{tabular}{l|cccc cccc cccc |c}
VLM &
\rotatebox{90}{CIFAR10}~~ &
\rotatebox{90}{CIFAR100}~~ &
\rotatebox{90}{Caltech101}~~ &
\rotatebox{90}{StanfordCars}~~ &
\rotatebox{90}{Flowers102}~~ &
\rotatebox{90}{Food101}~~ &
\rotatebox{90}{SUN397}~~ &
\rotatebox{90}{DTD}~ &
\rotatebox{90}{Aircrafts}~~ &
\rotatebox{90}{OxfordPets}~~ &
\rotatebox{90}{EuroSAT}~~ & 
\rotatebox{90}{\textbf{ImageNet}}~~ &
\rotatebox{90}{\textbf{Average}}~~ \\

\noalign{\smallskip}
\hline
\noalign{\smallskip}
 - & 
 63.9&
31.8&
62.3&
2.4&
48.8&
40.6&
43.3&
14.8&
3.8&
23.3&
15.1&
29.8& 31.7 \\
CLIP-ViT-L/14~\cite{radford2021learning} & 
66.5&
\textbf{40.8}&
70.6&
\textbf{5.6}&
57.7&
\textbf{55.2}&
\textbf{51.9}&
25.4&
4.9&
\textbf{48.0}&
25.9&
\textbf{41.2}&
41.1\\
$\text{FILIP}_\text{large}$\cite{yao2021filip}  & 
\textbf{71.2}& 
39.6& 
\textbf{73.3}& 
4.7& 
\textbf{57.9}& 
53.0& 
49.4& 
\textbf{26.9}& 
\textbf{6.0}& 
42.5& 
\textbf{34.2}& 
39.5& 
\textbf{41.5}\\
\noalign{\smallskip}
\hline
\end{tabular}
}
\end{table*}

\begin{table*}[t!]
\caption{Ablation studies of different vocabulary to retrieve visual concepts by using the VLM pre-trained on YFCC26M. The experiments are conducted on CLIP with ViT-B/16 backbone and image augmentation. The first model is pre-trained without visual concepts and is the VLM used to retrieve visual concepts for the next two models.
The results are Top-1 accuracy(\%) of zero-shot image classification on 12 datasets. ``VC'' is the Visual Concept. All results apply prompt ensembling.
}
\label{tab:yfcc26m-pretrained-self-retrieve}
\centering
\resizebox{\textwidth}{!}{
\begin{tabular}{l|cccc cccc cccc |c}
VLM &
\rotatebox{90}{CIFAR10}~~ &
\rotatebox{90}{CIFAR100}~~ &
\rotatebox{90}{Caltech101}~~ &
\rotatebox{90}{StanfordCars}~~ &
\rotatebox{90}{Flowers102}~~ &
\rotatebox{90}{Food101}~~ &
\rotatebox{90}{SUN397}~~ &
\rotatebox{90}{DTD}~ &
\rotatebox{90}{Aircrafts}~~ &
\rotatebox{90}{OxfordPets}~~ &
\rotatebox{90}{EuroSAT}~~ & 
\rotatebox{90}{\textbf{ImageNet}}~~ &
\rotatebox{90}{\textbf{Average}}~~ \\

\noalign{\smallskip}
\hline
\noalign{\smallskip}
CLIP & 75.3
 & 42.4  & 69.5 & 3.9 & 54.8
 & 51.1
 & 46.6 & 18.6 & 3.9 & 21.7
 & 20.5 & 39.2 & 37.3 \\
~w/ YFCC VC & 
\textbf{82.6} &
47.1&
\textbf{74.2}&
\textbf{6.0}&
57.8&
58.3&
\textbf{49.9}&
20.4&
\textbf{6.2}&
30.6&
26.0&
\textbf{42.7}&
\textbf{41.8} \\
~w/ Ensemble VC & 
78.3&
\textbf{48.5}&
70.1&
4.9&
\textbf{58.7}&
\textbf{58.5}&
47.6&
\textbf{22.4}&
5.6&
\textbf{31.7}&
\textbf{28.9}&
42.5&
41.5 \\
\noalign{\smallskip}
\hline
\end{tabular}
}
\end{table*}

\subsection{Comparison of Different Visual Concept Vocabulary}
The visual concept vocabulary we used in our pre-training is collected from several corpora. Table \ref{tab:different-concepts-corpus} shows the comparison when pre-training with the visual concepts from different corpora. Experimental results demonstrate that the wider range the visual concept vocabulary covers, the better performance achieves. In Fig. \ref{fig:nltk-code-example}, we show a code example of constructing the visual concept vocabulary from WordNet of NLTK.

\begin{table*}[t!]
\caption{Ablation studies of various corpus to construct the visual concept vocabulary. The results are Top-1 accuracy(\%) of zero-shot image classification on 12 datasets. All models are pre-trained on YFCC26M and follow the architecture of CLIP with ViT-B/32 backbone, MAE decoder and 50\% masking ratio. OWT denotes the OpenWebText corpus\cite{gokaslan2019openwebtext}.
Only use single prompt: `a photo of a [classname]'.
}
\label{tab:different-concepts-corpus}
\centering
\resizebox{\textwidth}{!}{
\begin{tabular}{l|c|cccc cccc cccc |c}
Source & 
\# Noun &
\rotatebox{90}{CIFAR10}~~ &
\rotatebox{90}{CIFAR100}~~ &
\rotatebox{90}{Caltech101}~~ &
\rotatebox{90}{StanfordCars}~~ &
\rotatebox{90}{Flowers102}~~ &
\rotatebox{90}{Food101}~~ &
\rotatebox{90}{SUN397}~~ &
\rotatebox{90}{DTD}~ &
\rotatebox{90}{Aircrafts}~~ &
\rotatebox{90}{OxfordPets}~~ &
\rotatebox{90}{EuroSAT}~~ & 
\rotatebox{90}{\textbf{ImageNet}}~~ &
\rotatebox{90}{\textbf{Average}}~~ \\
\noalign{\smallskip}
\hline
\noalign{\smallskip}
 - & - & 
 63.9&
31.8&
62.3&
2.4&
48.8&
40.6&
43.3&
14.8&
3.8&
23.3&
15.1&
29.8& 31.7 \\
OWT & 10000 & 
67.3& 
40.9& 
71.3& 
4.1& 
50.8& 
49.3& 
48.7& 
19.7& 
3.5& 
32.3& 
27.6& 
36.6& 
37.7 \\
YFCC & 10000 & 
63.8& 
39.5& 
65.8& 
\textbf{4.7}& 
55.5& 
50.6& 
47.7& 
19.8& 
4.6& 
39.9& 
29.6& 
36.2& 
38.1 \\
NLTK & 67176 & 
69.2& 
38.7& 
67.3& 
4.6& 
53.3& 
51.7& 
\textbf{49.6}& 
19.6& 
5.8& 
37.2& 
30.4& 
37.6& 
38.8\\
N-gram & 100000 & 
\textbf{72.0} &
\textbf{41.3}&
71.9&
3.8&
56.7&
50.7&
48.6&
20.2&
5.0&
39.4&
20.3&
36.5&   
38.9 \\
Ensemble & 151912 & 
71.2& 
39.6& 
\textbf{73.3}& 
\textbf{4.7}& 
\textbf{57.9}& 
\textbf{53.0}& 
49.4& 
\textbf{26.9}& 
\textbf{6.0}& 
\textbf{42.5}& 
\textbf{34.2}& 
\textbf{39.5}& 
\textbf{41.5}\\
\noalign{\smallskip}
\hline
\end{tabular}
}
\end{table*}

\section{Discussion of Different Text Encoder}
\label{sec:discussion-text-encoder}
% the pre-train model influences compare to CLIP: parameters and pre-trained internalized.
The original text encoder of CLIP~\cite{radford2021learning} is a GPT\cite{radford2019language}-like architecture that applies the causal mask to perform autoregressive manner, while we employ BART\cite{lewis2020bart} as the text encoder. Table \ref{tab:pre-trained-language-model} shows the comparison of different language models on the zero-shot classification of ImageNet. We conduct the experiments on CLIP. Without a pre-trained language model, training with BART performs worse than the original CLIP model. Training with a pre-trained BART will bring an improvement of 2.6\%.
% And pre-training with BERT achieves better zero-shot performance on ImageNet than pre-training with BART no matter with pre-trained initialized or not. 
% Note that, in this experiment, BERT doesn't add the masked language modeling task but BART adds the text filling task.
\begin{table}[t]
\caption{Comparison of different language model architecture with ViT-B/32. All models are pre-trained on YFCC26M and follow the architecture of CLIP with ViT-B/32 backbone, MAE decoder and 50\% masking ratio. The models with BART receive the masked text in the text encoder. ``Top-1 Acc'' means the top-1 accuracy of zero-shot classification on ImageNet. Only use single prompt: `a photo of a [classname]'.}
\label{tab:pre-trained-language-model}
\centering
\resizebox{\linewidth}{!}{
\begin{tabular}{l|cccc}
\hline
\noalign{\smallskip} 
Language Model & Params & Pre-trained & Top-1 Acc \\
\noalign{\smallskip}
\hline
\noalign{\smallskip} 
CLIP-GPT~\cite{radford2021learning} & 165M & \xmark & 30.1 \\
%  BERT~\cite{devlin2018bert} & 211M & \xmark & 32.1 \\
%  BERT~\cite{devlin2018bert} & 211M & \cmark & \textbf{34.6} \\ 
BART~\cite{lewis2020bart}  & 241M & \xmark & 29.0 \\
BART~\cite{lewis2020bart}  & 241M & \cmark & \textbf{32.7} \\
\noalign{\smallskip} 
\hline
\end{tabular}
}
\end{table}
While existing works, \emph{e.g.}, ImageBERT~\cite{qi2020imagebert}, ALBEF~\cite{li2021align}, employ BERT \cite{devlin2018bert} as the text encoder backbone because of its good Natural Language Understanding(NLU) performance, NLIP apples BART\cite{lewis2020bart} under the consideration of better Natural Language Generation(NLG). BART is a sequence-to-sequence architecture that combines Bidirectional Transformers (\emph{i.e.}, BERT) and Auto-Regressive Transformers~(\emph{i.e.}, GPT). 
With BART, NLIP is able to pre-training with the cross-modal contrastive learning (aligning the masked image with completed text) and the language modeling (reconstructing the masked text of text encoder), simultaneously. With BERT, pre-training with masked language modeling is harmful to cross-modal contrastive learning because the masked input of image and text might lose the key information and lead to misalignment. 
\begin{figure}
    \centering
\begin{lstlisting}[language=python]
import pickle
from nltk.corpus import wordnet as wn

nouns = {x.name().split('.', 1)[0] 
         for x in wn.all_synsets('n')}
nouns = [' '.join(n.split('_')) for n in nouns]
pickle.dump(nouns, open('nltk_wordnet_noun.pkl', 'wb'))
\end{lstlisting}
    \caption{A code example to get a clean visual concept vocabulary from NLTK.}
    \label{fig:nltk-code-example}
    % \vspace{-7mm}
\end{figure}
\section{More Discussion}
\label{sec:more-discussion}

\begin{table*}[t!] 
\caption{
Influence of pre-training CLIP with COCO caption. CLIP w/ coco means pre-training on YFCC26M and COCO caption simultaneously. All results apply prompt ensemble.
% ``Top-1 Acc'' means the top-1 accuracy of zero-shot classification on ImageNet. All models are with ViT-B/16.
% Ablation studies of different corpus to retrieve visual concepts by using the VLM pre-trained on YFCC26M. The experiments are conducted on CLIP with ViT-B/16 backbone and image augmentation. The first model is the VLM used to retrieve visual concepts.
% The results are Top-1 accuracy(\%) of zero-shot image classification on 12 datasets. VCC is the Visual Concept. All results apply prompt ensemble.
}
\label{tab:coco-info-influence}
\centering
% \resizebox{\textwidth}{!}{
\begin{tabular}{l|cccc cccc cccc |c}
Model &
\rotatebox{90}{CIFAR10}~~ &
\rotatebox{90}{CIFAR100}~~ &
\rotatebox{90}{Caltech101}~~ &
\rotatebox{90}{StanfordCars}~~ &
\rotatebox{90}{Flowers102}~~ &
\rotatebox{90}{Food101}~~ &
\rotatebox{90}{SUN397}~~ &
\rotatebox{90}{DTD}~ &
\rotatebox{90}{Aircrafts}~~ &
\rotatebox{90}{OxfordPets}~~ &
\rotatebox{90}{EuroSAT}~~ & 
\rotatebox{90}{\textbf{ImageNet}}~~ &
\rotatebox{90}{\textbf{Average}}~~ \\

\noalign{\smallskip}
\hline
\noalign{\smallskip}
 CLIP & 75.3
 & 42.4  & \textbf{69.5} & 3.9 & \textbf{54.8}
 & \textbf{51.1}
 & \textbf{46.6} & \textbf{18.6} & \textbf{3.9} & 21.7
 & 20.5 & \textbf{39.2} & \textbf{37.3}  \\
CLIP w/ COCO &
\textbf{78.6}&
\textbf{43.3}&
65.8&
\textbf{4.0}&
52.3&
48.2&
43.7&
18.1&
2.8&
\textbf{22.6}&
\textbf{30.2}&
37.2&
37.2 \\
\noalign{\smallskip}
\hline
\end{tabular}
% }
\end{table*}
\paragraph{Influence of COCO caption} 
% At the \emph{captioning} stage, we fine-tune the captioner on COCO caption for generating better synthetic text, which is indirectly\hjh{indirectly?} using COCO caption data within our pre-training.
Due to NLIP using COCO caption at \emph{captioning} stage, for the sake of fair comparison, we directly pre-train CLIP on YFCC26M and COCO caption simultaneously. The experimental results, shown in Table \ref{tab:coco-info-influence}, demonstrate that directly pre-training on YFCC26M and COCO Caption doesn't achieve significant improvement but slightly drop the performance on specific datasets, \emph{e.g.}, ImageNet, Caltech101, while NLIP with Noise Completion can achieve 0.5\% improvement on ImageNet and nearly 3\% improvement on Caltech101. This is because the Noise Completion generates finer descriptions of images on YFCC26M that makes full use of the model's caption generation capability and denoises the dataset. And COCO caption with limited categories and 113K images has little impact on pre-training.

% \paragraph{Noise Harmonization} The scale set to 1 will bring too strong regularization and drop the performance about 3.6\% while a 0.5 scale will bring 0.9\% improvements. Fig \ref{fig:after-warmup-distribution}
\begin{table}
\centering
\caption{Ablation study of the visual encoder-decoder module. Mask Ratio represents the masking proportion of the input image. ``Top-1 Acc.'' means the top-1 accuracy of  zero-shot classification on ImageNet. PE indicates positional embedding. The first row represents the baseline without masking the input image tokens.}
% \resizebox{0.95\linewidth}{!}{{
% \footnotesize
\begin{tabular}{cc|c}
\hline
\noalign{\smallskip} 
Mask Ratio(\%) & PE & Top-1 Acc.(\%)\\
\noalign{\smallskip}
\hline
\noalign{\smallskip} 
-   & \cmark & 31.3 \\
75  & \xmark & 26.5 \\
75  & \cmark & 27.0 \\
50  & \cmark & 30.1 \\
25  & \cmark & 30.4 \\
\noalign{\smallskip} 
\hline
\end{tabular}
%}} % end of resizebox
\label{tab:mae}
\end{table}
\paragraph{Abaltion of Visual Encoder-Decoder} 
Table~\ref{tab:mae} studies the effect of key factors in the visual encoder-decoder of NLIP, including the masking ratio and positional embedding, on ImageNet zero-shot classification task. Experiments are conducted with ViT-B/32 as the visual encoder and without using data augmentation. 
% , which is an effective indicator for the quality of the learned vision-language representations. 
Besides, we adopt a hand-crafted prompt \textit{`a photo of a \{category\}'} for evaluation. 
Results in Table~\ref{tab:mae} shows that adding the positional embedding brings the gain of 0.5\% on top-1 accuracy (row 2 vs. row 3) in zero-shot classification. As we can see, a high masking proportion (e.g., 50\% masking ratio) of the input image still reserves 30.1\% zero-shot top-1 accuracy on ImageNet which just slightly drops compared to the baseline 31.3\%. 
Moreover, experiments show that the high masking ratio can speed up training, which is consistent with \citet{he2021masked}.
\begin{figure*}[t!]
\centering
\includegraphics[width=\linewidth]{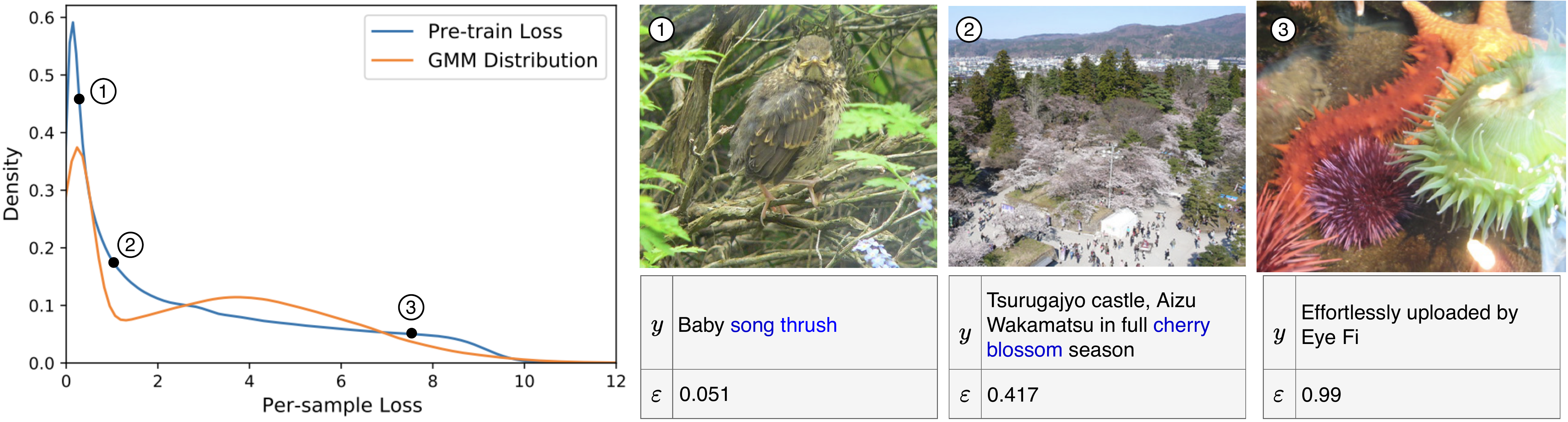}
\caption{Illustration of per-sample loss distribution of the final epoch and the distribution of GMM predictions in Eq. (7).
The left side shows three examples with different distribution locations and predicted noise probabilities. The right side shows three image-text pairs at corresponding locations.}
\label{fig:loss-distribution}
\end{figure*}
\paragraph{Benefit of Masked Image Encoder} 
The masked image encoder randomly masked image patches to save memory and accelerate the training processing.
We conduct an additional experiment to demonstrate the effectiveness of the masked image encoder. As Table~\ref{tab:masked-acceleratioin-largest-bs} shows, with 50\% masking ratio, we process 448 more samples and save 27\% pre-training time on single node with 8 NVIDIA V100. Besides, the masked image encoder can achieve 1\% improvement on zero-shot classification of ImageNet. Note that gradient checkpoint is not utilized for this comparison.
\begin{table}
\caption{Benefit of Masked Image Encoder. 
``Top-1 Acc'' means the zero-shot performance on ImageNet. The experiment is conducted on single node with 8 NVIDIA V100 GPU with ViT-B/16. Only use single prompt: `a photo of a [classname]'.}
\label{tab:masked-acceleratioin-largest-bs}
\centering
\resizebox{\linewidth}{!}{
\begin{tabular}{cccc}
\hline
\noalign{\smallskip} 
Model & Batch Size  & Hour/Epoch & Top-1 Acc \\
\noalign{\smallskip}
\hline
\noalign{\smallskip} 
 w/o     masking   & 1152 & 5.5  & 33.4 \\  
 w/ 50\% masking   & 1600 & 4    & \textbf{34.4} \\  
\noalign{\smallskip} 
\hline
\end{tabular}
}
\end{table}
% \paragraph{NoCaps Results} Noval Object Captioning is a extended image captioning task to test the model's capability to describe the noval objects from the Open Image datset\cite{}. We use the NoCaps \cite{} dataset to evaluate the model performance when meeting out of-domain objects. Note that we break the 

% \paragraph{More Linear probing results.} Table \ref{tab:linear-classification-B32} shows the linear probing image classification results of FILIP and NLIP on 11 datasets. All models are pre-trained with ViT-B/32 backbone. NLIP achieves an absolute 9.1\% average improvement over 12 datasets compared to FILIP.

\noindent\textbf{Effect of Noise Harmonization.}
Fig. \ref{fig:loss-distribution} shows the per-sample loss distribution at the final epoch of the noise-aware pre-training stage and the distribution predicted by the two-component GMM. 
% It is shown that the most of data is treated as correctly matching pairs which acquire small loss before fitting with our .
Three image-text examples located on different points of the per-sample loss distribution are shown in the right side of Fig.~\ref{fig:loss-distribution}. 
We can observe that along with the increased per-sample loss,
the predicted noise probability is increasing and the relevance between the image and text is decreasing. 
It verifies the reliability of our noise-harmonization scheme by dividing the unlabeled image-text pairs into clean and noisy sets implicitly with the memorization effect of the cross-model transformer.
\begin{figure}[t!]
\centering
\includegraphics[width=\linewidth]{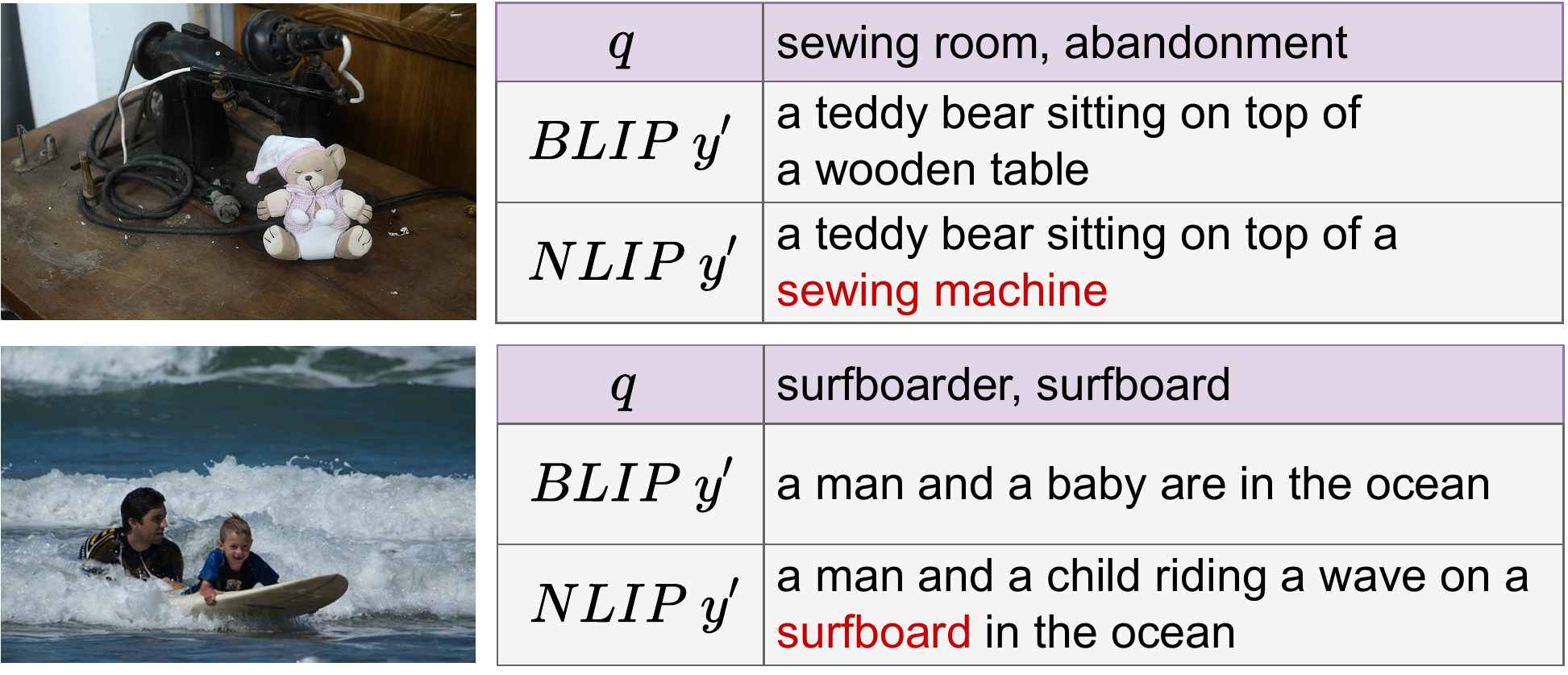}
\caption{
Qualitative captioning comparison of BLIP and NLIP. With the visual concepts $Q$ as the auxiliary input, NLIP can better complete the caption with more detailed information of existing objects in the image.}
\label{fig:compare_blip}
\end{figure}

\begin{figure}[t!]
\centering
\includegraphics[width=\linewidth]{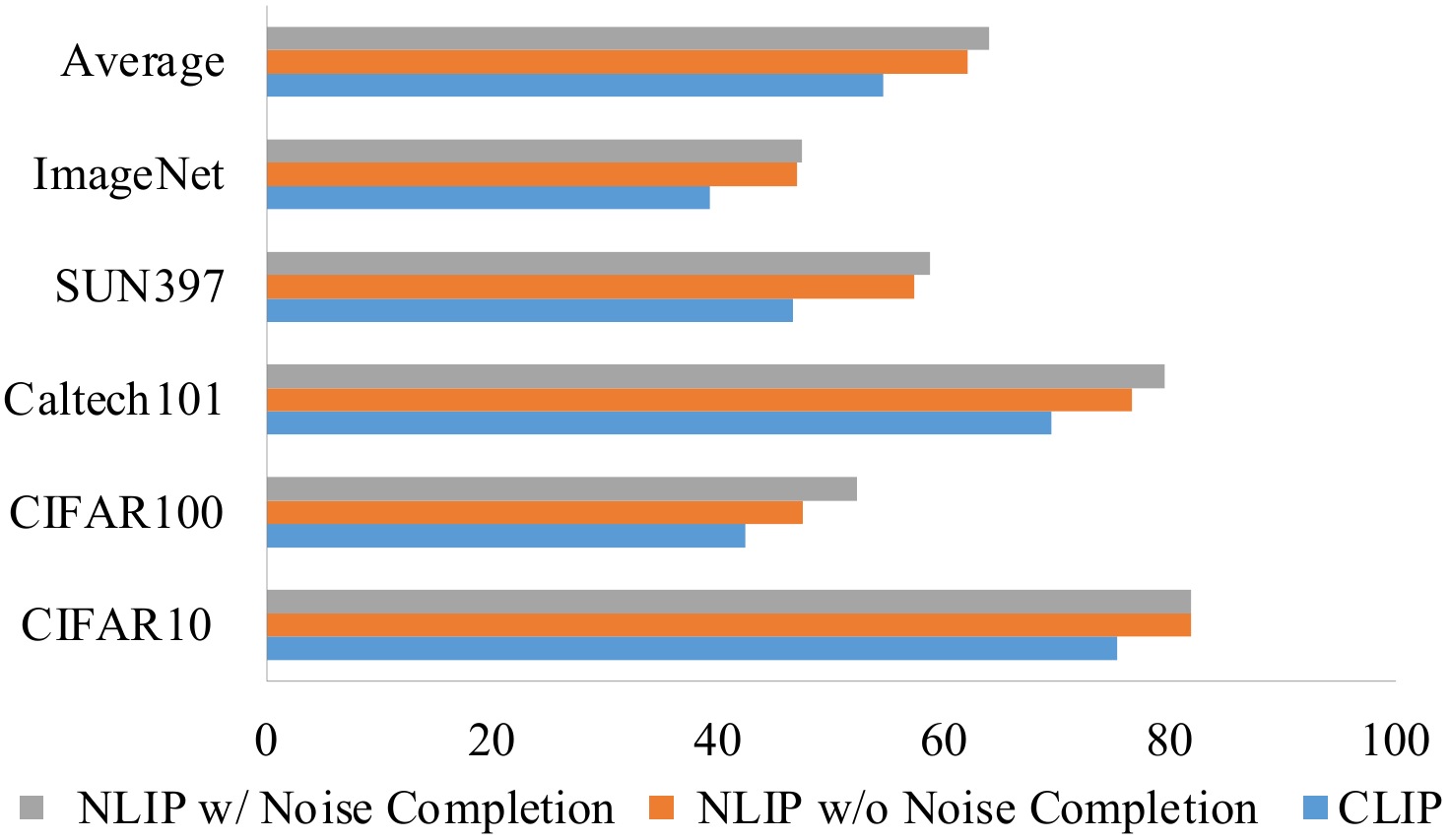}
\caption{
Ablation study of noise completion scheme on zero-shot image classification on five datasets. The first row is the average performance of these five datasets.}
\label{fig:noise-completion-ablation}
\quad
\end{figure}
\begin{figure*}[t!]
\centering
\includegraphics[ width=\linewidth]{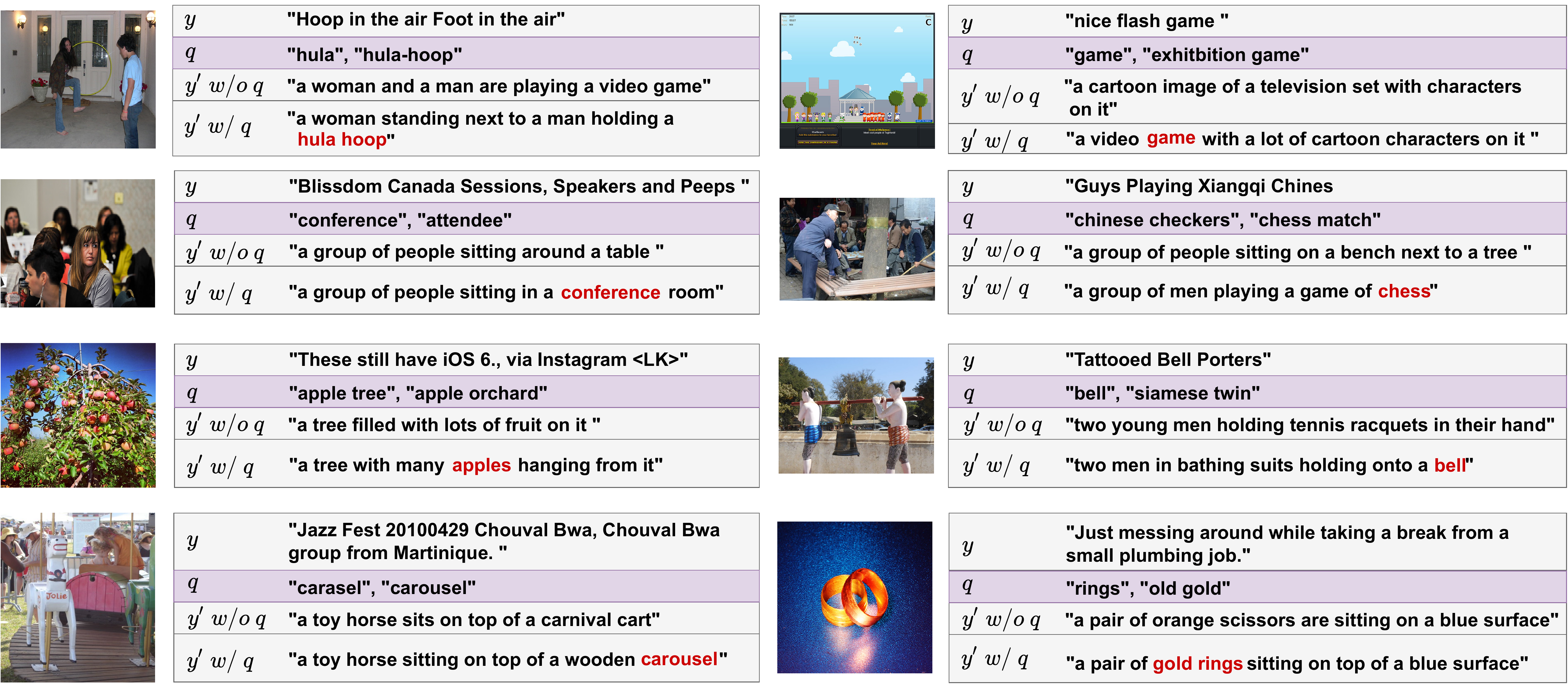}
\caption{
More examples about the synthetic captions generated by NLIP. Note that $y$, $y'$ and $q$ denote the web text, synthetic text and the visual concepts, respectively.
Better generation performance can be achieved with visual concepts as the auxiliary input. $\langle LK\rangle$ means the dropped web link.} 
\label{fig:more-synthetic-caption-for-yfcc}
\end{figure*}

\noindent\textbf{Effect of Noise Completion.} 
Fig.~\ref{fig:noise-completion-ablation} reports the zero-shot classification results of our NLIP with or without using the noise completion scheme on five different datasets. We can observe the noise completion scheme helps boost the performance by over 1.9\% on average accuracy. Furthermore, we illustrate some examples of enriched synthetic captions of YFCC via noise completion scheme in 
Fig.~\ref{fig:more-synthetic-caption-for-yfcc}. We can observe that the synthetic captions generated by NLIP show more concrete meanings and contain more semantic information compared to the original web texts.

\noindent\textbf{Fair comparison with BLIP} We pre-train BLIP on YFCC26M to provide fair comparison with our method since BLIP uses a lot more data (129M vs 26M) than us.
Note that we compare NLIP w/o noise completion with BLIP w/o using synthetic captions. 
Both using Image-Text Contrastive loss and being evaluated on Flickr30k, NLIP achieves 75.5 and  49.7 at R@1 on Image-to-text (I2T) and Text-to-image (T2I) retrieval while BLIP achieves 68.3 and 31.0 at R@1 on I2T and T2I retrieval.

\noindent\textbf{Retrieval on Conceptual Captions (CC3M)} We evaluate on a random-sampled subset of CC3M (with 1000 image-text pairs, similar to following Flickr's karpathy test set). As shown in Table \ref{tab:cc1k-retrieval}, NLIP surpasses FILIP and CLIP by 6.8 and 10.3 on R@1 of image-to-text retrieval.

\begin{table}
\caption{Results of zero-shot image-to-text(I2T) and text-to-image(T2I) retrieval on a 1k subset of CC12M.
}
\label{tab:cc1k-retrieval}
\centering
\begin{tabular}{l|cccc}
\hline
\noalign{\smallskip} 
\multirow{2}{*}{ Model } & \multicolumn{2}{c}{I2T} & \multicolumn{2}{c}{T2I} \\
\noalign{\smallskip} 
     & R@1 & R@10 & R@1  & R@10 \\
\noalign{\smallskip}
\hline
\noalign{\smallskip} 
CLIP  & 14.8 & 39.3  & 15.1 & 41.0  \\
FILIP & 18.3 & 44.7  & 19.4 & 46.9  \\
NLIP  & 25.1 & 53.5  & 24.7 & 54.1  \\
\noalign{\smallskip} 
\hline
\end{tabular}
\end{table}

\end{document}